 \documentclass[final,5p,times]{elsarticle}
\usepackage{amssymb}
\usepackage{amssymb} 
\usepackage{algorithm} 
\usepackage{caption}
\usepackage{hyperref}
\usepackage{subcaption}
\usepackage{tabularx}
\usepackage{array}
\usepackage{color}
\usepackage{graphicx}
\usepackage{booktabs}
\usepackage{float}
\usepackage{url}
\usepackage{xcolor}
\usepackage{placeins}
\usepackage{float}
\usepackage{setspace}

\journal{.}

\begin{document}

\begin{frontmatter}



\title{Detoxify: A framework for abusive text transformation   using LLMs}

 %: Groq, Gemini, GPT-4o and DeepSeek

\author[inst2,inst1]{Rohitash Chandra}

\author[inst1,inst2]{Jiyong Choi}
%\author[inst3]{Tanuj}

\author[inst2]{ Jayesh Sonawane}
\affiliation[inst2]{organization={Transitional Artificial Intelligence Research Group, School of Mathematics and Statistics, UNSW}, 
         state={Sydney},
       country={Australia}}

\affiliation[inst2]{organization={Centre for Artificial Intelligence and Innovation, Pingla Institute}, 
           state={Sydney},
            country={Australia}}
\begin{abstract}  

Although Large Language Models (LLMs) have demonstrated significant advancements in natural language processing tasks, their effectiveness in the classification and transformation of abusive text into non-abusive versions remains an area for exploration. In this study, we present Detoxify: a framework that employs LLMs to transform abusive text (tweets and reviews) containing hate speech and profanity into non-abusive text while retaining the original intent. We evaluate the performance of \textcolor{black}{four state-of-the-art LLMs, such as Gemini, GPT-4o, DeekSeek and  Groq}, on their ability to identify abusive text. We aim to transform and obtain a text that is clean of abusive and inappropriate content, but maintains a similar level of sentiment and semantics, i.e. the transformed text needs to maintain its message. Afterwards, we evaluate the raw and transformed datasets with sentiment analysis and semantic analysis. \textcolor{black}{Our results show Groq provides vastly different results when compared with other LLMs. We have identified similarities between GPT-4o and DeepSeek.  Groq stood out as the most distinct, as it often restructured sentences with excessive positive phrasing, with the original context lost or altered.}  
 
\end{abstract}


\end{frontmatter}


\section{Introduction}

%https://aclanthology.org/2024.emnlp-main.1066.pdf

%[paragraph about social media and abuse]
Social media platforms such as  Instagram, Facebook, Twitter (X), and TikTok enable users to create and share creative, entertaining and educational multimedia content \cite{carr2015social,kaplan2010users}. Social media enables users to freely comment, message, and express their opinions that can be displayed to the public or restricted groups, depending on the users' choice. The increasing popularity of social media and the ease of access to these online services have led to cyberbullying \cite{cyberbullying}, harassment, toxicity, and other types of abuse across   platforms \cite{
pater2016characterizations,van2014means}.
Although social media is relatively new, it is also evolving with strategies such as short-form content \cite{wu2021relationship} on platforms such as TikTok, Instagram (Reels), and YouTube (Shorts), targeting the younger generation, with the rise of \textit{influencers} \cite{hudders2021commercialization}. Since short-form content mostly consists of video with audio,  toxicity and abusive content are typically unmonitored, which brings danger to teens and young children who are also users of these platforms \cite{richards2015impact}. 

Social media platforms face the challenge of identifying and mitigating harmful and abusive material while ensuring that freedom of expression is not limited. 
This task is critical for maintaining healthy online environments where users can express themselves without fear of harassment \cite{zampieri2020semeval}. As a result, developing and applying advanced language models has become central to improving the accuracy and efficiency of sentiment classification, particularly in distinguishing between abusive content \cite{olkhovikov2022maximality}.  
An increase in toxicity can often be due to anonymity, lack of responsibility, and echo chambers in social media where users group themselves, sharing common beliefs and narratives to reinforce their old beliefs \cite{doi:10.1073/pnas.2023301118}. 
Although platforms have implemented reporting systems and moderation \cite{gongane2022detection}, these tools have shown limited effects, leaving much abuse and toxicity prevalent on social media and leaving users in the dark \cite{myers2018censored}.

Natural Language Processing (NLP) \cite{manning1999} and Machine Learning methods have been used to detect and moderate abuse \cite{
jahan2023systematic,alrashidi2022review}.  Early efforts in abuse detection primarily focused on rule-based approaches and keyword matching to identify abuse,  which struggled with the complexity and variability of human language. Therefore,  sophisticated techniques were necessary to handle the nuanced and context-dependent nature of abusive comments \cite{schmidt2017survey}.  NLP \cite{
otter2020survey,torfi2020natural} powered by deep learning have significantly transformed text processing. Sentiment analysis is an NLP method  \cite{wankhade2022survey} that plays a critical role in market research and social media content moderation. Sentiment analysis provides an analysis of emotions and opinions across vast volumes of textual data. 

Large Language Models (LLMs) \cite{brown2020language,naveed2024comprehensiveoverviewlargelanguage}  are NLP methods that utilise large deep learning models with a massive corpus of data and pre-trained data that can be refined with domain-specific data for NLP tasks, including classification, text generation, text summarisation, and question answering \cite{10.1371/journal.pone.0255615}. Notable advancements in discussion-based LLMs include GPT-4o \cite{openai2024gpt4o}, Gemini \cite{gemini2024family}, and Groq \cite{groq2026api}. Although GPT-4o and Gemini are recognised for their comprehensive understanding of context and nuance, Groq distinguishes itself with its focus on high-speed processing, open-source nature,  and scalability \cite{sridhar2022estimating}. However, as LLMS become complex, it is crucial to evaluate their performance in real-world scenarios to ensure they can effectively handle the nuances of human language.

In the past decade, much attention has been given to generative artificial intelligence models \cite{bandi2023power} that are used to create audio-visual content, digital art and animation \cite{gozalo2023chatGPT}, making them prominent in social media. LLM-based sentiment analysis for product reviews has been done \cite{amirifar2023nlp}, but not much in creating synthetic reviews. However, LLMs are becoming prominent in monitoring abusive content \cite{jaremko2025revisiting}. 
LLMs have been under debate and controversy for academic essays and use in writing research papers \cite{
buruk2023academic,megawati2023role,lazebnik2024detecting}. 

Furthermore, LLMs can also be used for language translation tasks, and a study evaluated them for Indian language translation (Hindi, Sanskrit, Telugu), where LLMs have shown better performance than Google Translate and GPT \cite{chandra2025evaluation}.
 The power of text generation and generative language models can be utilised to generate and transform abusive reviews into non-abusive ones. NLP methods have been used to generate product descriptions based on product reviews \cite{novgorodov2019generating}.
Understanding how LLMs handle and transform abusive comments can inform the development of more robust tools for online safety, ultimately supporting healthier digital environments \cite{rothman2022transformers}. 
Although there has been an increasing sophistication of methods used to detect abusive text, there is a lack of comparative analysis of emerging LLMs, such as the Gemini and Groq models. The power of text generation with LLMs can be utilised to generate and transform abusive reviews into non-abusive ones.

In this study, we present Detoxify: a framework that employs s to transform abusive text featuring hate speech and swear words (tweets and reviews)  into non-abusive text, where the message and emotion are retained, i.e. the semantics and the    sentiment. We evaluate the performance of two state-of-the-art LLMs on their ability to identify abusive text and transform it.  \textcolor{black}{Our key research goals involve 1) abuse detection, 2) abuse transformation, and 3) evaluation of preservation of sentiment and meaning of the transformed text in comparison to the original text.}
 We utilise two datasets of abusive text obtained from private forums and social media that feature swear words, personal attacks, and racist remarks. We then use Gemini-1.5-flash, Groq (llama3-8b-8192), GPT-4o and DeepSeek-V3 to transform them so that we obtain a text that is clean of abusive and inappropriate content but maintains a similar level of sentiment and semantics. We use  BERT-based models to evaluate the raw and transformed datasets using sentiment analysis and semantic analysis. \textcolor{black}{The novelty of this study is in our framework for automated text transformation and evaluation.  We provide insights into the respective strengths and weaknesses of the models, offering a deeper understanding of how each model can contribute to safer and more respectful online communities. }

\label{sec:sample1}

\section{Background}

\subsection{Deep learning and  LLMs}

Prominent deep learning models are based on Convolutional Neural Networks (CNNs)\cite{oshea2015introductionconvolutionalneuralnetworks} and Recurrent Neural Networks (RNNs), such as Long Short-Term Memory (LSTM) networks \cite{hochreiter1997long,van2020review} and Bidirectional LSTM \cite{GRAVES2005602}  models.  \textcolor{black}{The  Transformer model is an advanced deep learning model that features an attention mechanism that offers superior performance for long-range dependencies and faster training via parallel processing\cite{vaswani2023attentionneed}.}
 Deep learning models can learn hierarchical representations of text and have been better at capturing context and sequence information, making them more effective in detecting abusive content.    RNNs have been leveraged to model the temporal dependencies in text, allowing for a better understanding of the context in which potentially abusive language was used \cite{li2021data}. CNNs were designed for visual data but have also been effective in extracting relevant features from text through convolutional layers, enhancing the detection of abusive comments even in short and unstructured text data \cite{zhang2018detecting}. 

Pre-trained LLMs, such as BERT, capture bidirectional context in understanding language nuances and have set new benchmarks in various NLP tasks, including the detection of toxic and abusive language \cite{Saleh31122023} and sentiment analysis \cite{10.1371/journal.pone.0255615}. Studies demonstrated that fine-tuning BERT on datasets labelled for abusive content can significantly improve the accuracy and robustness of detection models \cite{mishra2019tackling}. Furthermore, transfer learning, where models on large general-purpose corpora are fine-tuned on smaller, domain-specific datasets to improve performance on tasks such as abusive language detection \cite{yan2020re}. Additionally, researchers have investigated the challenges of dataset bias and the importance of diverse and representative training data \cite{sap2019risk}. Some studies have highlighted the issue of models disproportionately flagging particular dialects or demographics as abusive, which has prompted efforts to create more balanced and fair datasets \cite{dixon2018measuring}. The use of adversarial training, where models are trained to recognise subtle, context-dependent abusive content while minimising bias, has been explored to address these concerns \cite{guarcello2018josephson}. Fine-tuned BERT models have been used in different domains; for example, MedBERT \cite{rasmy2020medbertpretrainedcontextualizedembeddings} is based on structured diagnosis data from electronic health records and is suitable for medical decision-making applications. Furthermore, Hate-BERT \cite{caselli-etal-2021-hatebert}  has been pre-trained with an abusive language dataset and applied to a wide range of problems, including the analysis of abuse in Hollywood movie dialogues over seven decades \cite{chandra2025longitudinalabusesentimentanalysis}.
Therefore, we can utilise models such as Hate-BERT \cite{caselli-etal-2021-hatebert} to analyse the level of abuse in a given text in our study.

\subsection{LLMs in operation}

Gemini \cite{gemini2024family} is an LLM developed by Google DeepMind and \textcolor{black}{known for its deep reasoning and multimodal capabilities, as well as creative tasks such as image creation, text generation, and literature review. } A report \cite{geminiteam2024gemini15unlockingmultimodal}  showed that Gemini 1.5-Flash outperformed GPT-4-Turbo in terms of latency and speed across English, Japanese, Chinese and French queries. In terms of the audio haystack, Gemini 1.5 Flash successfully identified 98.7\% of instances, whereas GPT-4Turbo achieved an accuracy of 94.5\%.

\textcolor{black}{Groq \cite{grok2026} is an LLM powered by Llama3 \cite{grattafiori2024llama3herdmodels}, which employs Latency Processing Units (LPUs) designed specifically to minimise latency and maximise performance for LLM inference.} Unlike Gemini, Groq-based models can only receive text input; however, Groq allows for lower latency and reduced computational costs.  GPT-4o \cite{openai2024gpt4o} is an LLM developed by OpenAI, originally released as ChatGPT. It is known for coding and problem-solving skills similar to Gemini; however, it only accepts up to 128,000 tokens, whereas Google's Gemini model accommodates up to 1 million tokens. DeepSeek\cite{DeepSeekai2025DeepSeekr1incentivizingreasoningcapability}  is a new LLM developed in China that rivals state-of-the-art technology with low-cost development. DeepSeek is free and open-source and gained much attention in January 2025 due to its high performance and low cost-effectiveness. The DeepSeek-V3 technical report \cite{DeepSeekai2025DeepSeekv3technicalreport} shows that DeepSeek performed better than GPT in all coding and maths benchmarks (for models larger than 67B) and most English tasks.
Furthermore, the Llama-3 (developed by Facebook/Meta) report\cite{grattafiori2024llama3herdmodels} shows that Llama3-8b underperforms when compared to GPT in every benchmark, as expected considering the difference in size of the models.  

\subsection{Detoxification}
\textcolor{black}{Recent work on text detoxification has framed the abusive text problem as a style-transfer or controlled rewriting task in which toxic or unsafe language is transformed into non-toxic text while preserving semantic meaning and fluency \cite{dale2021detoxification,hallinan2023marco}. Early detoxification approaches relied on unsupervised style transfer and paraphrase-guided generation to balance toxicity reduction with semantic preservation.   Dale et al.~\cite{dale2021detoxification} proposed paraphrase-guided detoxification using LLMs and demonstrated that content preservation remains one of the central challenges in safe rewriting.   MaRCo \cite{hallinan2023marco} is a software system designed to control revision strategies using expert and anti-expert language models to better preserve contextual meaning, while removing subtle toxicity and microaggressions.}

\textcolor{black}{The broader literature on paraphrasing and stylistic rewriting further emphasises the importance of semantic consistency during transformation. Context-aware rewriting approaches have shown that sentence-level detoxification can produce generic or semantically distorted outputs when discourse context is ignored \cite{yerukola2023contextual}. Human evaluations in contextual rewriting studies consistently demonstrated that users prefer rewrites that preserve communicative intent and emotional tone rather than merely minimising toxicity scores \cite{yerukola2023contextual}. These findings align with recent detoxification datasets that explicitly evaluate sentiment preservation and emotional fidelity during rewriting \cite{wang2025toxirewritecn}. Further work concerns refusal behaviour and safety alignment in LLMs, as safety-tuned models may over-refuse benign or research-oriented prompts containing toxic language, even when the intended task is detoxification or analysis rather than harmful generation. Im et al.~\cite{im2026falserefusal} systematically studied false refusal behaviour in hate-speech detoxification and found that refusal rates increase for prompts involving protected groups, political references, and semantically toxic wording.  Khondaker et al. \cite{khondaker2024detoxllm} presented a detosixty framework that used LLMs to transform text with explanations to promote transparency and trustworthiness.}

\textcolor{black}{Human evaluation remains a critical component in assessing detoxification quality because automated metrics alone cannot capture semantic fidelity, fluency, and contextual appropriateness. Therefore, evaluation studies have been done on detoxification systems using a combination of toxicity reduction, semantic similarity, and fluency measures alongside human judgments \cite{dale2021detoxification,hallinan2023marco}. Manual evaluation protocols often assess whether rewrites preserve the original intent while removing offensive content, with annotators rating fluency, meaning preservation, and stylistic safety independently. It has been shown that human evaluators frequently prioritise contextual appropriateness and communicative coherence over strict lexical similarity \cite{yerukola2023contextual}, hence essential for measuring the quality and faithfulness of safe rewriting systems.}

\section{Methodology}

\subsection{Abuse datasets}
 
We use a dataset extracted at the Indian Institute of Technology (IIT) Guwahati, consisting of over 160,000 entries from various platforms, including Reddit, 4chan and Twitter via web-scraping \footnote{\url{https://github.com/pinglainstitute/LLM-reviewtransformation/tree/main/Dataset}}. We refer to this as the IIT-abuse dataset, where each entry has an abuse indicator, where 1 means abusive text and 0 means non-abusive text.

We also use the Twitter (X) data  \cite{user_profiling_and_abusive_language_detection_dataset_543}, which consists of 4265 tweets collected from 52 users. The tweets contain samples of five different categories of abusive messages, namely religion, NSFW ( Not Suitable For Work), racism, discrimination and non-abusive. We note that NSFW is used to mark links to content, videos, or website pages that may contain graphic violence, pornography, profanity, and other forms of inappropriate content. 
The account names of users have to be anonymised so as not to violate the policy about privacy.

\textcolor {black}{Ethical considerations are central to the data lifecycle, given that the datasets include abusive user-generated content, web-scraped material, and potentially privacy-sensitive information. All data used has a strict minimisation principle, ensuring only necessary content is retained for analysis. Web-scraped sources have been restricted to publicly available material, with explicit avoidance of paywalled, private, or access-controlled data without permission. User-generated content has been anonymised at ingestion by removing direct and indirect identifiers, including usernames, locations, contact details, and any metadata that could enable re-identification.    Ethical oversight, including institutional review where applicable, our results are stored in a GitHub repository given at the end of the paper, guided by privacy regulations and responsible AI principles.}

\subsection{BERT-based models for abuse detection and sentiment analysis }
%### Methodology

We employ BERT-based pre-trained NLP models as a foundational component in evaluating the effectiveness of the four models for sentiment analysis. BERT's ability to fine-tune specific tasks with minimal additional training data makes it particularly effective in sentiment analysis \cite{10.1371/journal.pone.0255615}, semantic analysis \cite{9715095} and abuse detection \cite{chandra2025longitudinalabusesentimentanalysis}.

We fine-tune the BERT-based model on the sentiment analysis dataset (SenWave data \cite{yang2020senwavemonitoringglobalsentiments}) to establish a baseline performance for multi-label sentiment classification. 
 The SenWave dataset features Tweets from March 1 2020, to May
15th 2020 and contains 10,000 human-labelled Tweets, which enabled it to be utilised in various studies, particularly for refining pre-trained language models.  It  includes 10 different emotions (Optimistic, Thankful, Empathetic, Pessimistic, Anxious,
Sad, Annoyed, Denial, Official Report, Joking). For example, Chandra et al. \cite{9715095}
evaluated the Bhagavad Gita translation (Sanskrit to English)  BERT model based
on the SenWave dataset. The fine-tuning process involves adjusting the weights of the pre-trained BERT model to optimise it for the specific task of identifying abusive and non-abusive content. \textcolor{black}{In the SenWave dataset, the Official Report is data about COVID-19 official reports and is not useful for our study. We will not utilise this in the results.}

In case of abuse detection, we utilise Hate-BERT \cite{caselli-etal-2021-hatebert}, which has been pre-trained and refined using RAL-E, a dataset consisting of text from banned Subreddits for harmful and hateful content. HateBERT was trained on over 1.5 million messages consisting of over 40 million tokens, and a Masked Language Model (MLM) was applied. HateBERT was then tested against BERT on three datasets and consistently showed greater performance compared to BERT.

\subsection{Detoxify Framework}

\begin{figure*}
    \centering
    \includegraphics[width=1.0\linewidth]{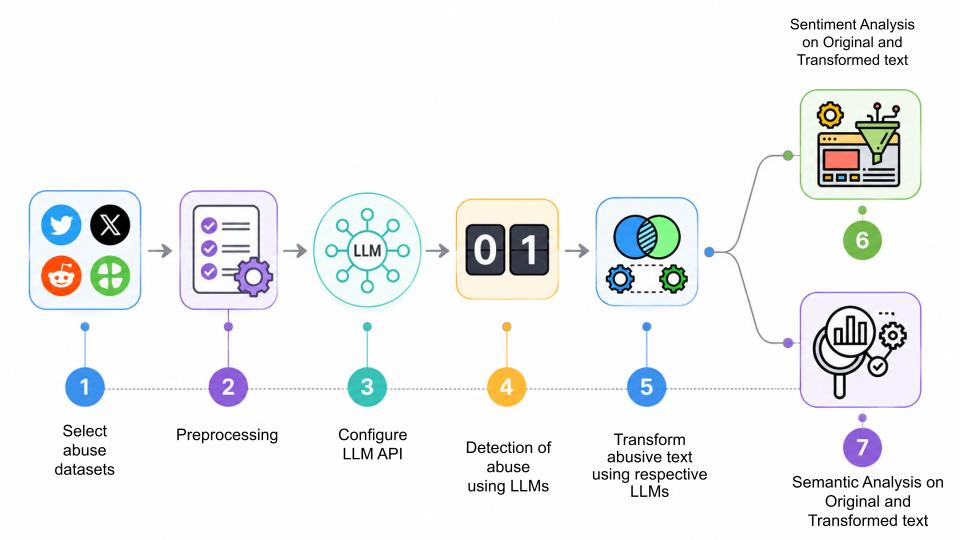}
    \caption{\textcolor{black}{Detoxify Framework for transforming abusive text into non-abusive text and conducting sentiment analysis using LLMs (Groq, Gemini, GPT-4o and DeepSeek).  }}
    \label{fig:framework}
\end{figure*}

%https://docs.google.com/presentation/d/1RhCNnlLZwHajkRmLRNWUUkCNPyGdOCZ3LV-vWJfXeVc/edit?usp=sharing

We present the Detoxify framework designed to transform abusive reviews into non-abusive ones while maintaining the original intention through semantics and sentiment   using state-of-the-art LLMs, including Groq, Gemini, GPT-4o and DeepSeek. Our framework (Figure \ref{fig:framework}) consists of seven stages: data acquisition, preprocessing, LLM API (Application Programming Interface)  configuration,  review transformation, and finally, sentiment and semantic analysis.

In Stage 1, we acquired a substantial dataset consisting of both abusive and non-abusive reviews. We extracted the datasets from popular user-generated platforms, including Reddit, 4chan, and Twitter, which provided a diverse range of content.  We pre-processed the datasets to standardise the text input, which involves tokenisation, lowercasing, and removing irrelevant symbols, ensuring that the text is in a suitable format for input into the models.

In Stage 2, we applied data preprocessing to ensure that the raw text data is ready for transformation and analysis. We developed a data cleaning pipeline to normalise the text data, starting with converting all reviews to lowercase for uniformity. We expanded the contractions, such as "won't" and "can't," to their full forms, making the text easier for models to interpret. We then removed unnecessary elements such as URLs and non-alphanumeric characters, which tend to introduce noise into the data. Tokenisation was performed to break the text into individual words, enabling the removal of stopwords; i.e. words that do not contribute meaningfully to sentiment or context, such as "the". We also apply lemmatisation to reduce words to their root forms, uniformly treating similar words (e.g., "running" and "run").  This preprocessing stage is essential for producing a clean and consistent dataset, priming the reviews for further stages of analysis.

In Stage 3, we set up the LLM Application Programmer Interface (API) for transforming abusive reviews. We configured the unique API keys and various other settings to optimise performance and accuracy. We also experimented with different API configurations and prompts to find the most effective settings for transforming reviews. 

In Stage 4, we test the respective LLMs on a subset of reviews extracted from our large dataset of mixed abusive and non-abusive content. Our goal is to evaluate each model's ability to predict whether a review is abusive or non-abusive. We explored the models' effectiveness in handling abusive language by experimenting with different API settings, such as modifying safety guidelines and adjusting language preferences. This early evaluation highlighted the strengths and limitations of each model, helping us identify areas where further fine-tuning could enhance performance for the larger dataset. These preliminary results laid the groundwork for the subsequent stages of our framework.
 We also implement bigram and trigram analysis to identify the most common sequence of words. We also compare the transformation rates between Gemini and Groq models.

In Stage 5, we utilise the respective LLMs for the transformation of abusive reviews into non-abusive and polite versions. We select a subset of reviews from the dataset and design a prompt to instruct the models to reframe the text in a polite and respectful tone.  

In Stage 6, we apply sentiment analysis to both the original and transformed reviews to understand how each of the models' text is classified, enabling us to identify the difference in sentiment distribution in LLM-transformed text.
 \textcolor{black}{We employ SenWave-BERT sentiment analysis model utilising the SenWave dataset \cite{yang2020senwavemonitoringglobalsentiments} as done in past studies \cite{singh_2025_14898138}.} The SenWave dataset includes over 105 million tweets and Weibo messages to evaluate the global sentiment during the COVID-19 pandemic. The 10 sentiment categories include: Optimistic, Thankful, Empathetic, Pessimistic, Anxious, Sad, Annoyed, Denial, Official Report, and Joking.  SenWave-BERT is a multi-label classification model, meaning a sentence can have multiple categories, e.g. Optimistic and Thankful.

In stage 7, we make use of the MP-Net semantic analysis model to generate embeddings and perform cosine similarity \cite{mikolov2013efficientestimationwordrepresentations} for each pairwise LLM-transformed text to compare the different LLMs with the original text. We also apply dimensionality reduction to visualise the embeddings in two dimensions to determine how semantically similar or different each model is to one another. 

Finally, we utilise evaluation metrics including accuracy, precision, recall, and F1-score. We focus on the model's ability to correctly identify abusive content, which is crucial for maintaining a healthy online environment. %\cite{papineni2002bleu}

\subsection{Technical details}

 We fine-tune the BERT-based model for sentiment analysis using supervised learning with the Adam optimiser \cite{kingma2014adam} with default key hyperparameters, such as learning rate (0.0001), batch size (32), and training epochs (max of 10). We used 10 output neurons for the major sentiments and ignored the Official Report sentiment.

To evaluate the performance of different language models in transforming offensive content, we implemented each model using its respective API. \textcolor{black}{The Groq model was initialised using the \texttt{Groq()} client with an API key, while the Gemini 1.5 Flash model was accessed via the \texttt{genai} Python library. Due to rate-limiting issues encountered with Gemini, we implemented a retry policy using Google’s built-in \texttt{Retry} class, configured with the following parameters: \texttt{initial=1.0}, \texttt{maximum=10.0}, \texttt{multiplier=2.0}, and \texttt{deadline=30.0}. We also turned off all safety settings, including hate speech, harassment, sexually explicit, and dangerous content, to help ensure the output was a transformed tweet and not a message regarding its policy.  We accessed GPT-4o and DeepSeek via the OpenAI Python library.} Although these models did not present issues related to rate limiting, they did not provide configurable options to disable safety filters, which limited direct control over content moderation.

\textcolor{black}{We presented each LLM the same prompt to ensure consistency across evaluations.}
In Stage 4, we used a specific LLM prompt for abuse detection: \begin{quote}
\textit{Transform the abusive tweet into a non-abusive one, i.e., into a more polite and respectful tweet while maintaining the overall meaning and context as much as possible. Only return the transformed tweet. Do not write anything apart from the transformed tweet. Below is the tweet:}
\end{quote}

In the case of abuse transformation in Stage 5, we used an LLM prompt: 
\begin{quote}
\textcolor{black}{\textit{"Transform the abusive tweet into a non-abusive one, i.e., into a more polite and respectful tweet while maintaining the overall meaning and context as much as possible. Only return the transformed tweet. Do not write anything apart from the transformed tweet. Below is the tweet:"}}
\end{quote}

\textcolor{black}{The LLM outputs were temperature-controlled. A fixed temperature of 1.0 was used across all LLMs to maintain comparability and reduce variability arising from sampling behaviour and verbosity differences. We controlled for verbosity bias by normalising the generated texts before similarity computation. Specifically, we applied consistent preprocessing and evaluated semantic similarity using embedding-based metrics on normalised sentence representations rather than raw token overlap. Additionally, we verified that the observed trends remained consistent after length normalisation, indicating that the higher similarity scores were not solely driven by longer Groq outputs. We note that no token truncation was done, as the average tweet length is very small compared to the context limits of the LLMs used. Due to API limitations, we only used 100 tweets  per model (Table \ref{tab:llm_pricing}).}

\begin{table*}[htbp!]
\centering
\small
\caption{Comparison of API Pricing and Free-Tier Availability Across Selected LLMs}
\label{tab:llm_pricing}
\begin{tabular}{|p{3cm}|p{2.5cm}|p{2cm}|p{5cm}|p{4.5cm}|}
\hline
\textbf{Model} & \textbf{Provider} & \textbf{Free Tier Availability} & \textbf{Free Tier Limits} & \textbf{Paid API Cost (Approx.)} \\
\hline

Llama 3.1 8B Instant 
& Groq 
& Yes 
& $\sim$14,400 requests/day, $\sim$500K tokens/day on free developer tier 
& Input: $\sim$\$0.05 / 1M tokens, Output: $\sim$\$0.08 / 1M tokens \\
\hline

Gemini 2.5 Flash 
& Google AI Studio 
& Yes 
& Free tier available with rate limits; limited daily prompts/tokens 
& Input: $\sim$\$0.30 / 1M tokens, Output: $\sim$\$2.50 / 1M tokens \\
\hline

GPT-4o 
& OpenAI API 
& Limited 
& Small trial credits for new users; no long-term free GPT-4o API tier 
& Input: $\sim$\$2.50 / 1M tokens, Output: $\sim$\$10 / 1M tokens \\
\hline

DeepSeek 
& DeepSeek 
& Partial 
& Free chat access; some platforms offer free API credits/tokens 
& Input: $\sim$\$0.27 / 1M tokens, Output: $\sim$\$0.42 / 1M tokens \\
\hline

\end{tabular}
\end{table*}

\section{Results}

\subsection{Data analysis} 

\begin{figure*}
    \centering
    \includegraphics[width=0.8\linewidth]{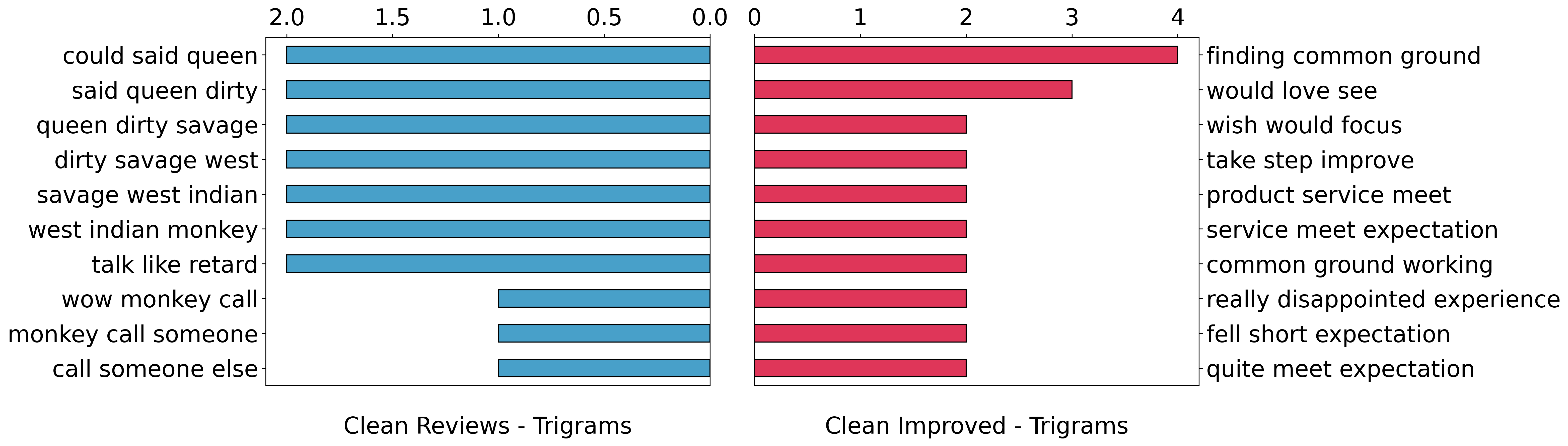}
    \caption{Trigram comparison  using Groq for showing original and transformed trigrams for the IIT dataset}
    \label{fig:trigrams}
\end{figure*}

Figure \ref{fig:trigrams} illustrates trigram frequency distributions for clean reviews and clean improved from the \textcolor{black}{IIT} dataset. The "Clean Reviews" trigrams contain highly offensive phrases like "west Indian monkey" and "talk like retard," which reflect severe instances of abusive language. This level of explicitness presents a unique challenge for NLP models as they must distinguish such terms from contextual or colloquial usage without allowing harmful language to pass undetected. The trigrams in the "Clean Improved" dataset show phrases transformed by the Groq LLM, such as "finding common ground" and "service meet expectation," indicating a shift toward constructive feedback. Terms like "would love see" and "take step improve" suggest positive engagement, which can foster better communication. This reframing capability indicates that Groq has the potential to contribute to more respectful discourse by reducing harmful language. 

We wish to clarify that the comparison was conducted using each model's default publicly available deployment settings. Contrary to the reviewer’s concern, the safety filters for Gemini were \textit{not disabled} during experimentation. All models were evaluated using their accessible API configurations without applying additional jailbreaks or bypass mechanisms.

\begin{table}[htbp!]
\centering
\caption{\textcolor{black}{Transformation Failure Analysis Across LLMs}}
\begin{tabular}{|l|c|c|}
\hline
\textbf{Model} & \textbf{Total Failures} & \textbf{Unique Failures} \\
\hline
Groq & 29 & 20 \\
Gemini & 15 & 5 \\
GPT & 4 & 0 \\
DeepSeek & 7 & 3 \\
\hline
Common failures & \multicolumn{2}{c|}{1} \\
\hline
\end{tabular}
\label{tab:failure_analysis}
\end{table}

\textcolor{black}{Out of the 500 abusive examples used for transformation experiments, Table \ref{tab:failure_analysis} shows the number of failed transformations for each LLM. These observations indicate that Gemini’s moderation mechanisms were active during experimentation, as evidenced by the non-zero number of rejected or failed transformations. We acknowledge that moderation policies differ across providers and can influence model behaviour. Therefore, we have revised the manuscript to explicitly frame the study as a comparison of \textit{deployed-system behaviour under provider-available settings}, rather than as a perfectly moderation-controlled benchmark.}

\textcolor{black}{In case of false positive cases (Figure 10), where the original text was already non-abusive, we manually reviewed selected cases and found that the LLMs still tended to rewrite the text into a more neutral or polished form rather than preserving it exactly \footnote{Transformed text available in GitHub repo: \url{https://github.com/pinglainstitute/LLM-reviewtransformation/tree/main/Results}}. For example, an original text \textit{“Can’t get over how popular I am these days”} was transformed into softer and more formal variants by multiple models despite containing no abusive content. This indicates a tendency toward over-sanitisation and stylistic normalisation even in non-abusive inputs. We therefore analysed semantic similarity and sentiment preservation to quantify such unnecessary transformations and identify potential meaning or tone drift. We provide more examples in Table 8.}

\subsection{Detection of abuse}

We first evaluate how effectively two selected LLMs identify abusive content in a given dataset. 
Table \ref{tab:accuracy_part4} provides a comparative analysis of the abusive word detection performance of  Gemini and Groq, across 22 batches of the  IIT-abuse dataset. We observe that Gemini generally reports higher abusive word counts compared to Groq in most batches, and it also obtains a higher mean count.  We note that Gemini had a higher F1 and IoU score than Groq, as shown in Table \ref{tab:compare-metrics}

\begin{table}[htbp]
\centering
\small
\caption{Comparison of Gemini and Groq using the mean across 22 batches of the IIT-abuse dataset.}
\label{tab:compare-metrics}
    \begin{tabular}{|c|c|c|}
    \hline
    \textbf{Metric} (mean) & \textbf{Gemini} & \textbf{Groq} \\
    \hline
    Precision & 0.338 & 0.305 \\
    \hline
    Recall & 0.786 & 0.804 \\
    \hline
    F1 & 0.442 & 0.419 \\
    \hline
    IoU & 0.314 & 0.291 \\
    \hline
    \end{tabular}
\end{table}

We further evaluate the performance of two LLMs, Gemini and Groq, on abuse in user reviews, with each model used to classify reviews as abusive (label=1) or non-abusive (label=0). \textcolor{black}{We note that the detection of abusive text in this case is essentially a binary classification problem.   The dataset consisted of 550 reviews, and was randomly divided into 22 batches of 25 reviews. This was also done to have fine-grained details using single error metrics on different segments of the data to provide a more comprehensive analysis.}  

\begin{figure}
    \centering
    \includegraphics[width=1.0\linewidth]{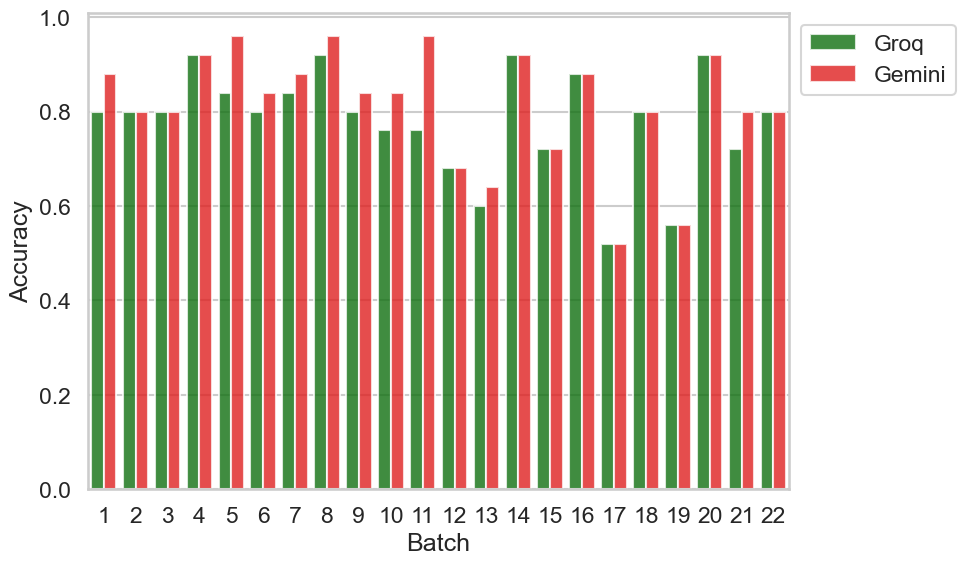}
    \caption{Batch-wise sentiment accuracy comparison of Groq and Gemini}
    \label{fig:batchwise}
\end{figure}

Table \ref{tab:accuracy_part4} presents the results where the mean accuracy achieved by \textcolor{black}{the Gemini model is 81.5 percent}. We examined the accuracy of each batch of 25 reviews to better understand the performance. The results indicated some variability in performance across different batches, as shown in Figure \ref{fig:batchwise}. The majority of batches achieved an accuracy of 80 percent, with some batches exceeding this baseline.

\begin{table}[htbp]
    \centering
    \small
    \caption{Sentiment Accuracy Comparison of Gemini and Groq by Batch Number for sentiment analysis using the  \textcolor{black}{IIT-abuse dataset.} We also report the mean and the standard deviation (Std).}
    \label{tab:accuracy_part4}
    \begin{tabular}{|c|c|c|}
        \hline
        \textbf{Batch No.} & \textbf{Gemini (\%)} & \textbf{Groq (\%)} \\
        \hline
        1  & 88 & 80 \\
        \hline
        2  & 80 & 80 \\
        \hline
        3  & 80 & 80 \\
        \hline
        4  & 92 & 92 \\
        \hline
        5  & 96 & 84 \\
        \hline
        6  & 84 & 80 \\
        \hline
        7  & 88 & 84 \\
        \hline
        8  & 96 & 92 \\
        \hline
        9  & 84 & 80 \\
        \hline
        10 & 84 & 76 \\
        \hline
        11 & 96 & 76 \\
        \hline
        12 & 68 & 68 \\
        \hline
        13 & 64 & 60 \\
        \hline
        14 & 92 & 92 \\
        \hline
        15 & 72 & 72 \\
        \hline
        16 & 88 & 88 \\
        \hline
        17 & 52 & 52 \\
        \hline
        18 & 80 & 80 \\
        \hline
        19 & 56 & 56 \\
        \hline
        20 & 92 & 92 \\
        \hline
        21 & 80 & 72 \\
        \hline
        22 & 80 & 80 \\
        \hline
       Mean  & 81.5 & 78.0 \\
        \hline
        Std & 12.1 & 11.0 \\
        \hline
    \end{tabular} 
\end{table}

The variability in accuracy across batches can be attributed to several factors, including the inherent variability in the review content and the potential influence of specific review characteristics on model performance. \textcolor{black}{Similarly, we employed  the Groq LLM for abuse detection, achieving an overall accuracy of  78 percent (Table \ref{tab:accuracy_part4}).}

\subsection{Transformation: comparison of  LLMs}

We begin by examining the success rates of transforming abusive reviews into non-abusive ones using the Gemini and Groq models. \textcolor{black}{The success rate measures the ability of the LLM to transform the review, where an unsuccessful transformation is the refusal of the LLM due to the extremely abusive nature of the text. LLMs have mechanisms to check and refuse requests, which is seen as an unsuccessful transformation. The dataset consists of 500 abusive reviews divided into 20 batches, each containing 25 reviews. Due to API restrictions, the transformations were performed in batches to ensure compliance with the limits of both models' APIs. This batch-wise processing also enabled us to systematically evaluate the consistency and reliability of each model across different review sets.}

\begin{figure}
    \centering
    \includegraphics[width=1.0\linewidth]{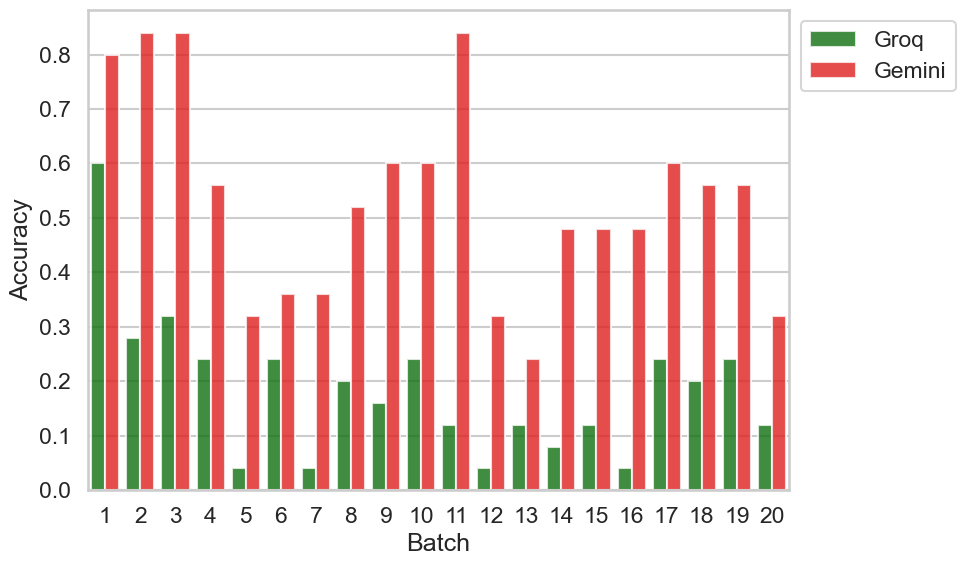}
    \caption{Batch-wise transformation accuracy comparison of Groq and Gemini}
    \label{fig:transformation_rate_part4}
\end{figure}

In Table \ref{tab:transfomation_rate}, we present the detailed statistics of the transformation success rates for each batch processed by the Gemini and Groq models. The success rates represent the percentage of reviews in each batch that were successfully transformed from abusive to non-abusive. The Gemini model consistently achieved higher success rates across the majority of batches compared to the Groq LLM (Figure \ref{fig:transformation_rate_part4}). For instance, in Batch 1, the Gemini model successfully transformed 80 percent of the reviews, whereas the Groq model only managed a 60 percent success rate. This trend continues across subsequent batches, with notable disparities such as Batch 11, where the Gemini model achieved an 84 percent success rate in contrast to Groq's 12 percent.

The disparity in success rates between the two models can be attributed to several factors, primarily related to API safety guidelines and settings. The Gemini model was configured with relaxed safety settings, allowing for a higher success rate in transforming abusive content. Specifically, the safety settings for the Gemini model were adjusted to block none of the harmful content categories.

The various \texttt{HarmBlockThreshold} settings available for the Gemini model are as follows:
\begin{itemize}
    \item \texttt{HARM\_BLOCK\_THRESHOLD\_UNSPECIFIED (0)}: Threshold is unspecified.
    \item \texttt{BLOCK\_LOW\_AND\_ABOVE (1)}: Content with negligible harm will be allowed.
    \item \texttt{BLOCK\_MEDIUM\_AND\_ABOVE (2)}: Content with negligible and low harm will be allowed.
    \item \texttt{BLOCK\_ONLY\_HIGH (3)}: Content with negligible, low, and medium harm will be allowed.
    \item \texttt{BLOCK\_NONE (4)}: All content will be allowed.
\end{itemize}

This configuration allowed Gemini to transform a greater number of abusive reviews into non-abusive ones without being constrained by stringent content moderation filters. In contrast, Groq did not offer such flexibility in safety settings, which likely contributed to its lower success rates. The lack of customisable safety settings in the Groq model means that it was more restricted in its ability to transform content, adhering to stricter guidelines that prevented certain transformations from occurring.

Thus, the higher transformation rates observed in the Gemini model, despite the inherent challenge of dealing with abusive content, can be partially attributed to the relaxed safety settings. Conversely, the Groq model’s adherence to stricter guidelines without the option to adjust safety settings resulted in a more conservative approach, leading to fewer successful transformations.

\begin{table}[htbp]
    \centering
    \small
    \caption{Transformation accuracy: comparison of Gemini and Groq for sentiment analysis using the  IIT-abuse dataset. We also report the mean and the standard deviation (Std).}
    \label{tab:transfomation_rate}
    \begin{tabular}{|c|c|c|}
        \hline
        \textbf{Batch No.} & \textbf{Gemini (\%)} & \textbf{Groq (\%)} \\
        \hline
        1  & 80 & 60 \\
        \hline
        2  & 84 & 28 \\
        \hline
        3  & 84 & 32 \\
        \hline
        4  & 56 & 24 \\
        \hline
        5  & 32 & 4 \\
        \hline
        6  & 36 & 24 \\
        \hline
        7  & 36 & 4 \\
        \hline
        8  & 52 & 20 \\
        \hline
        9  & 60 & 16 \\
        \hline
        10 & 60 & 24 \\
        \hline
        11 & 84 & 12 \\
        \hline
        12 & 32 & 4 \\
        \hline
        13 & 24 & 12 \\
        \hline
        14 & 48 & 8 \\
        \hline
        15 & 48 & 12 \\
        \hline
        16 & 48 & 4 \\
        \hline
        17 & 60 & 24 \\
        \hline
        18 & 56 & 20 \\
        \hline
        19 & 56 & 24 \\
        \hline
        20 & 32 & 12 \\
        \hline
        Mean & 53.4 & 18.4 \\
        \hline
        Std & 18.2 & 12.8 \\
        \hline
    \end{tabular} 
\end{table}

\subsection{Twitter (X) Data analysis}

We first clean the data, removing punctuation and lowering letters, then perform bigram and trigram analysis of 400 (100 for each abuse category).

In Figure \ref{fig:bigram_trigram_category}, we can see that common n-grams for Discriminatory abuse include "new mexico" and "new mexico compound" as well as "Muslim migrant" and "kill kill kill", targeting certain ethnic groups. For NSFW related abuse, "link bio" and "link bio *n*l" are the most common n-grams, followed by "bio *n*l" and "signup link bio". In terms of abusive racism, "see monkey" and "horrible little muslim" were the most common. For the bigrams, however, "monkey see", "horrible little", "muslim mayor" ", little muslim" and "look like" were just as common, which targets ethnic groups with discriminatory abuse, but in a more aggressive manner. Finally, common ngams related to religion included "sharia law" being the most frequent bigram by far, and "Prophet Muhammad pbuh" (peace be upon him), which directly relates to Islam.

\subsection{Transformation}

We perform transformation using the Groq, Gemini, GPT and DeepSeek LLMs. Table \ref{table:transformation_model} shows the aggregated sum of counts for each of the models. Overall, GPT-4o had the highest successful transformation, which implies it had the fewest restrictions when processing the abusive text, whereas Groq had the lowest number of successes, implying stricter guidelines.

\begin{table}[htbp]
    \centering
    \small
    \caption{Number of successful transformations by model (out of 400 Abusive examples)}
    \label{tab:transformations_success}
    \begin{tabular}{|c|c|c|}
        \hline
        \textbf{Model} & \textbf{Success} & \textbf{Fail}\\
        \hline
        GPT & 396 & 4\\
        \hline
        DeepSeek & 393 & 7\\
        \hline
        Gemini & 385 & 15\\
        \hline
        Groq & 371 & 29\\
        \hline
    \end{tabular}
    \label{table:transformation_model}
\end{table}

\begin{figure}
    \centering
    \includegraphics[width=1\linewidth]{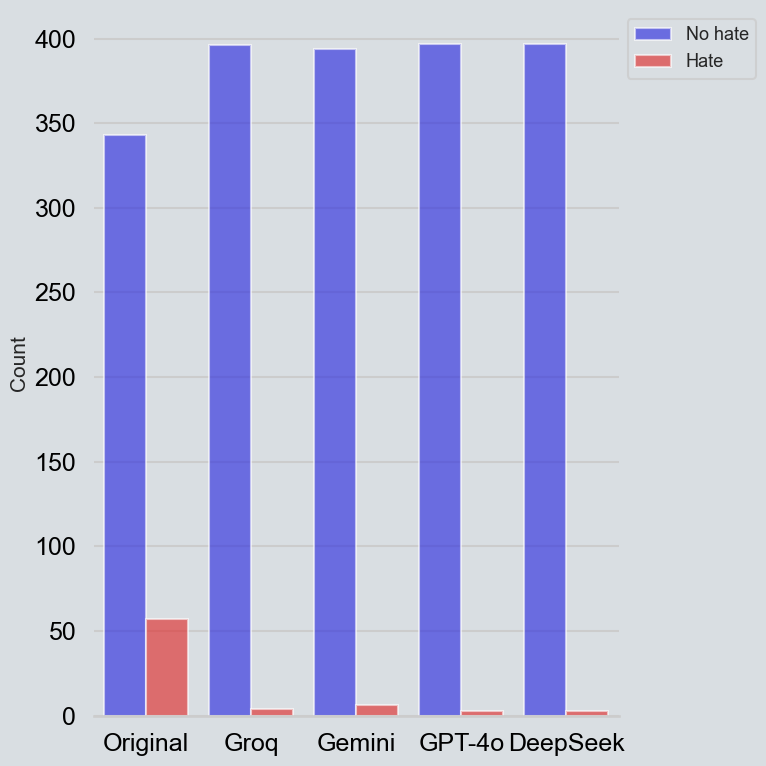}
    \caption{\textcolor{black}{Assessment of hateful words and phrases (Hate vs No Hate) captured by the LLM after transformation in comparison with the original text. }}
    \label{fig:hate_keyword_model}
\end{figure}

\begin{figure}
    \centering
    \includegraphics[width=1\linewidth]{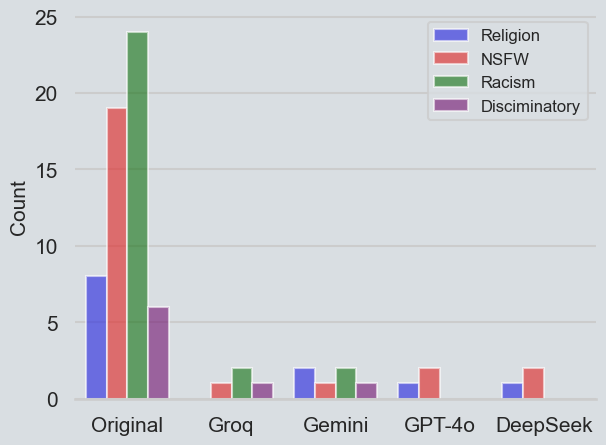}
    \caption {\textcolor{black}{Type of hateful words and phrases captured by the LLM after transformation in comparison with original text. }}
    \label{fig:hate_model_abusetype}
\end{figure}

\begin{figure}
    \centering
    \includegraphics[width=1\linewidth]{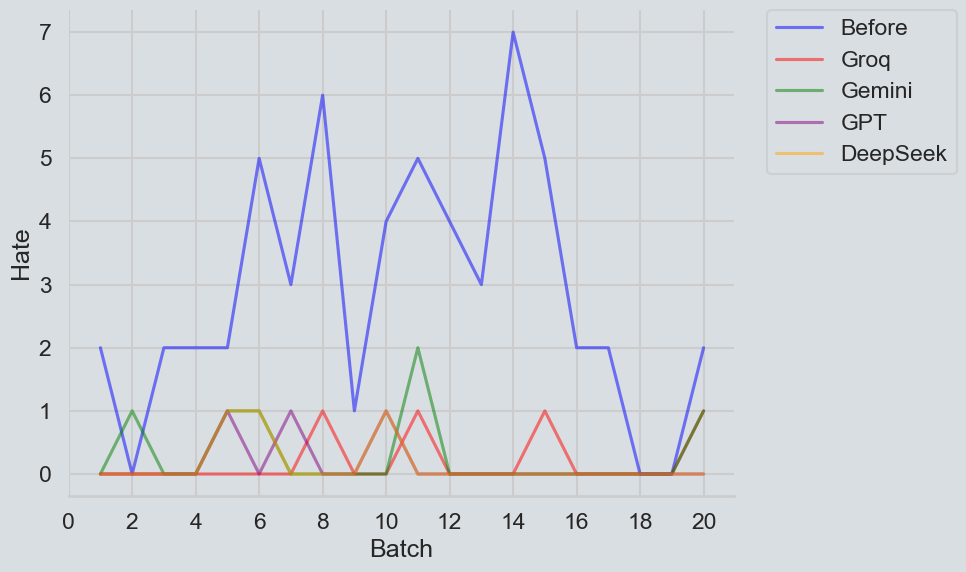}
    \caption{\textcolor{black}{Number of hateful words and phrases (Hate) per batch for the selected LLMs in comparison to the original text (Before).}}
    \label{fig:hate_lineplot}
\end{figure}

\subsection{Hate-Speech detection}

After the transformations, we  analyse the number of hateful words expressed by the LLM before and after the \textcolor{black}{transformations using a BERT-based hate speech model \cite{antypas-camacho-collados-2023-robust} trained from 13 hate-speech datasets and available via Cardiff-NLP \footnote{\url{https://huggingface.co/cardiffnlp/twitter-roberta-base-hate-multiclass-latest}}}. % \textcolor{black}{However, we encountered an error where the HateBERT model wrongly classified most tweets as hateful, which was not expected. This could have been partially because data was fine-tuned on Reddit posts in HateBERT, which can be quite long compared to tweets or due to technical issues during configuration.}

\begin{table}[htbp]
\centering
\caption{\textcolor{black}{Hate speech label transition analysis after LLM-based transformation.}}
\label{tab:hate_transition_analysis}

\begin{tabular}{lccc}
\hline
\textbf{Model} & \textbf{Removed Hate} & \textbf{Existing Hate} & \textbf{Hate Added} \\
\hline
Groq     & 79  & 0 & 0 \\
Gemini   & 78 & 1 & 0 \\
GPT-4o   & 79  & 0 & 0  \\
DeepSeek & 79  & 0 & 1 \\
\hline
\end{tabular}

\end{table}

\begin{figure}[htbp]
    \centering
    \includegraphics[width=1.05\linewidth]{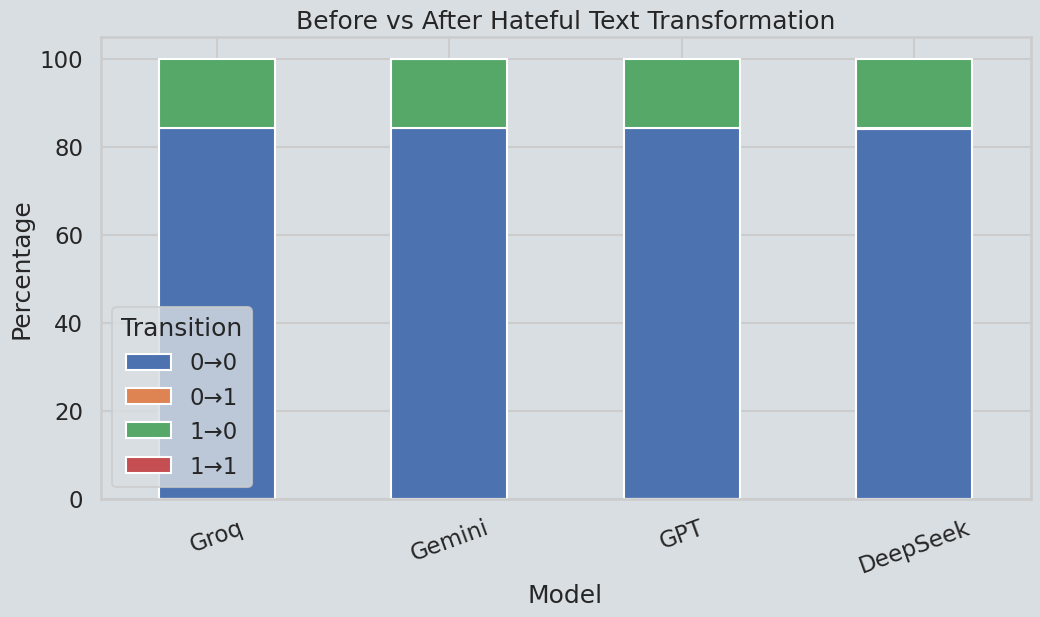}
    \caption{\textcolor{black}{Percentage distribution of hate label transitions before and after LLM-based transformation across different models.}}
    \label{fig:hate_transition_plot}
\end{figure}

\begin{figure}[htbp]
    \centering
    \includegraphics[width=1.05\linewidth]{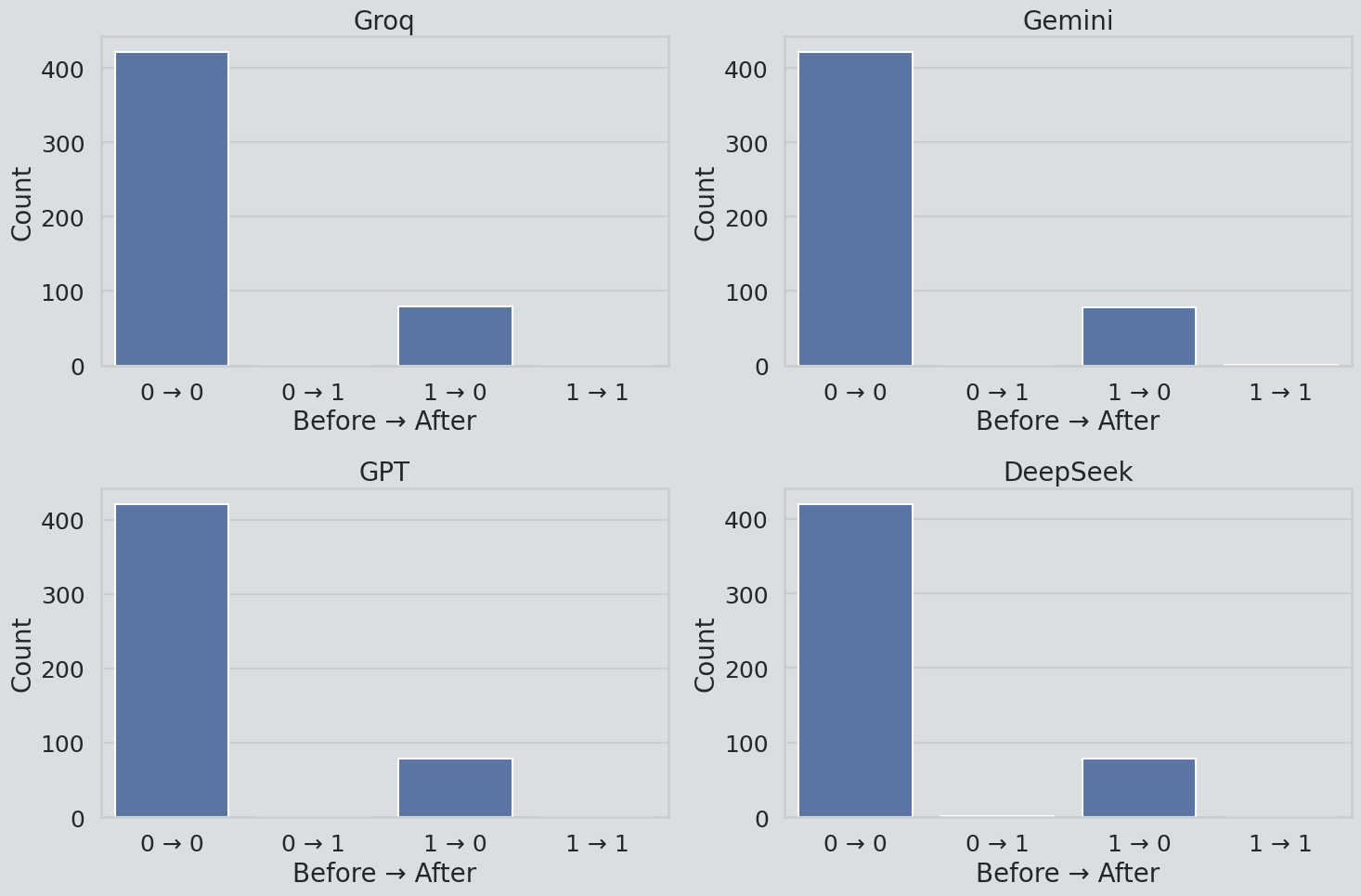}
    \caption{\textcolor{black}{Transition counts of HateBERT predictions before(0) and after(1) transformation for different LLMs.}}
    \label{fig:transition_analysis}
\end{figure}

\begin{figure}[htbp]
    \centering
    \includegraphics[width=1.05\linewidth]{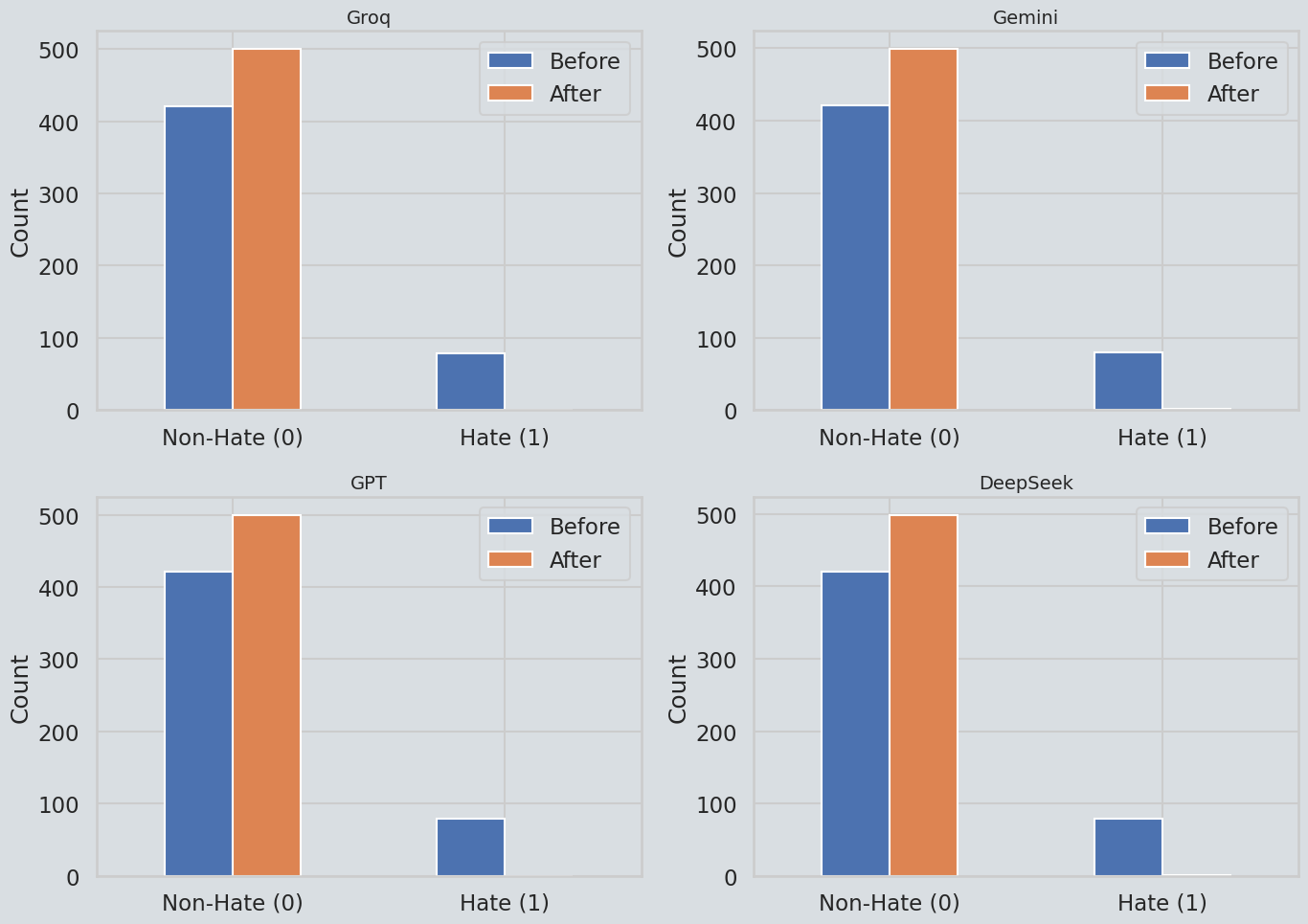}
    \caption{\textcolor{black}{Comparison of hate and non-hate classifications before and after text transformation using different LLMs, illustrating the conversion of hateful text into non-hateful text. }}
    \label{fig:transition_analysis}
\end{figure}
% \begin{figure}[htbp]
%     \centering

%     \begin{subfigure}[b]{0.5\linewidth}
%         \centering
%         \includegraphics[width=\linewidth]{Plots/trans_heatmap.png}
%         \caption{Groq}
%     \end{subfigure}
%     \hfill
%     \begin{subfigure}[b]{0.5\linewidth}
%         \centering
%         \includegraphics[width=\linewidth]{Plots/trans_heatmap_1.png}
%         \caption{Gemini}
%     \end{subfigure}

%     \begin{subfigure}[b]{0.5\linewidth}
%         \centering
%         \includegraphics[width=\linewidth]{Plots/trans_heatmap_2.png}
%         \caption{GPT-4o}
%     \end{subfigure}
%     \hfill
%     \begin{subfigure}[b]{0.5\linewidth}
%         \centering
%         \includegraphics[width=\linewidth]{Plots/trans_heatmap_3.png}
%         \caption{DeepSeek}
%     \end{subfigure}

%     \caption{\textcolor{black}{Percentage transition heatmaps of HateBERT predictions before and after transformation for different LLMs.}}
%     \label{fig:combined_transition_heatmaps}

% \end{figure}

\begin{figure}
    \centering
    \includegraphics[width=\linewidth]{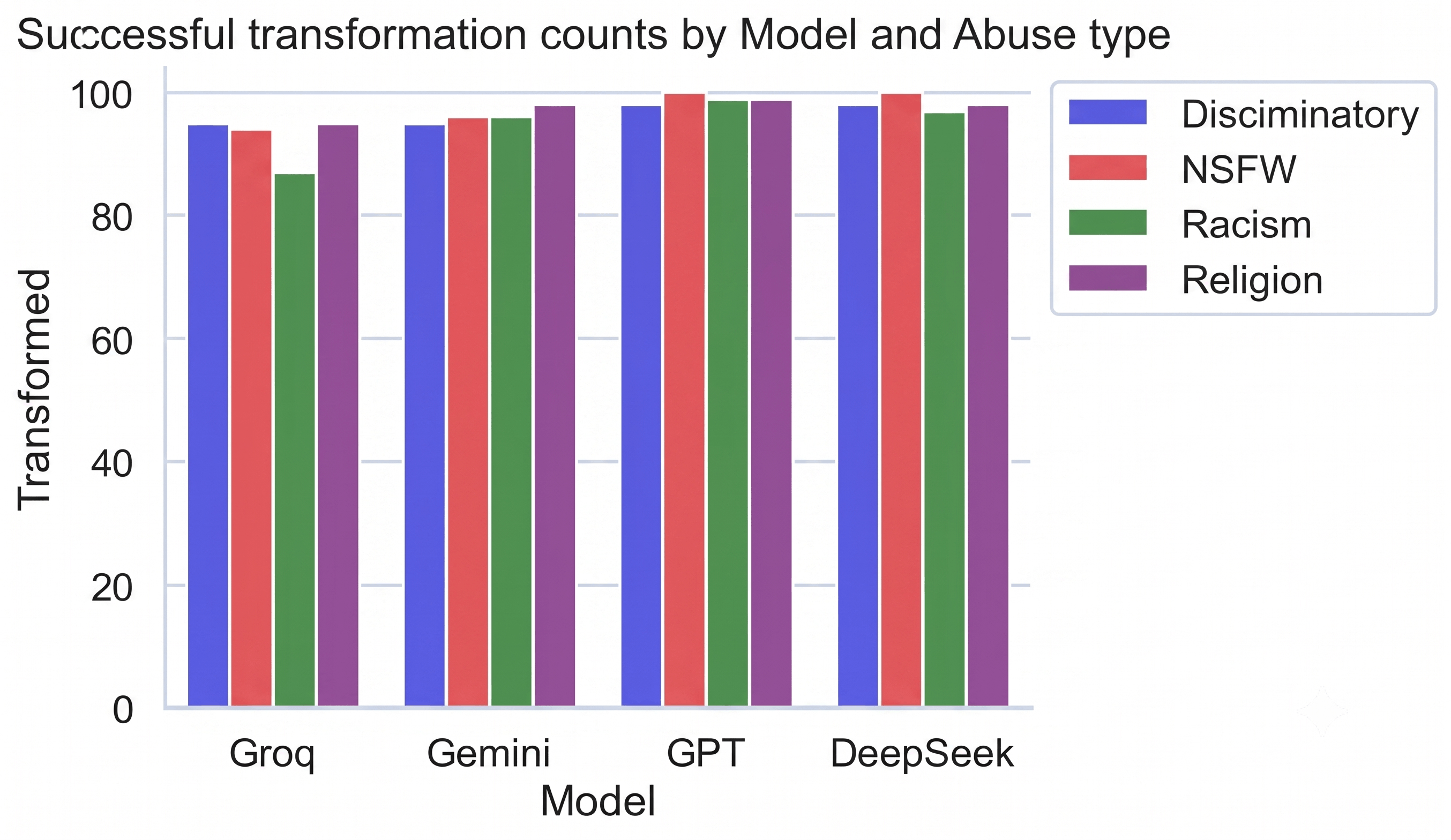}
    \caption{Successful transformation counts by Model and Abuse type}
    \label{fig:transformation_model_abuse}
\end{figure}

\begin{figure*}
    \centering
    \includegraphics[width=0.7\linewidth]{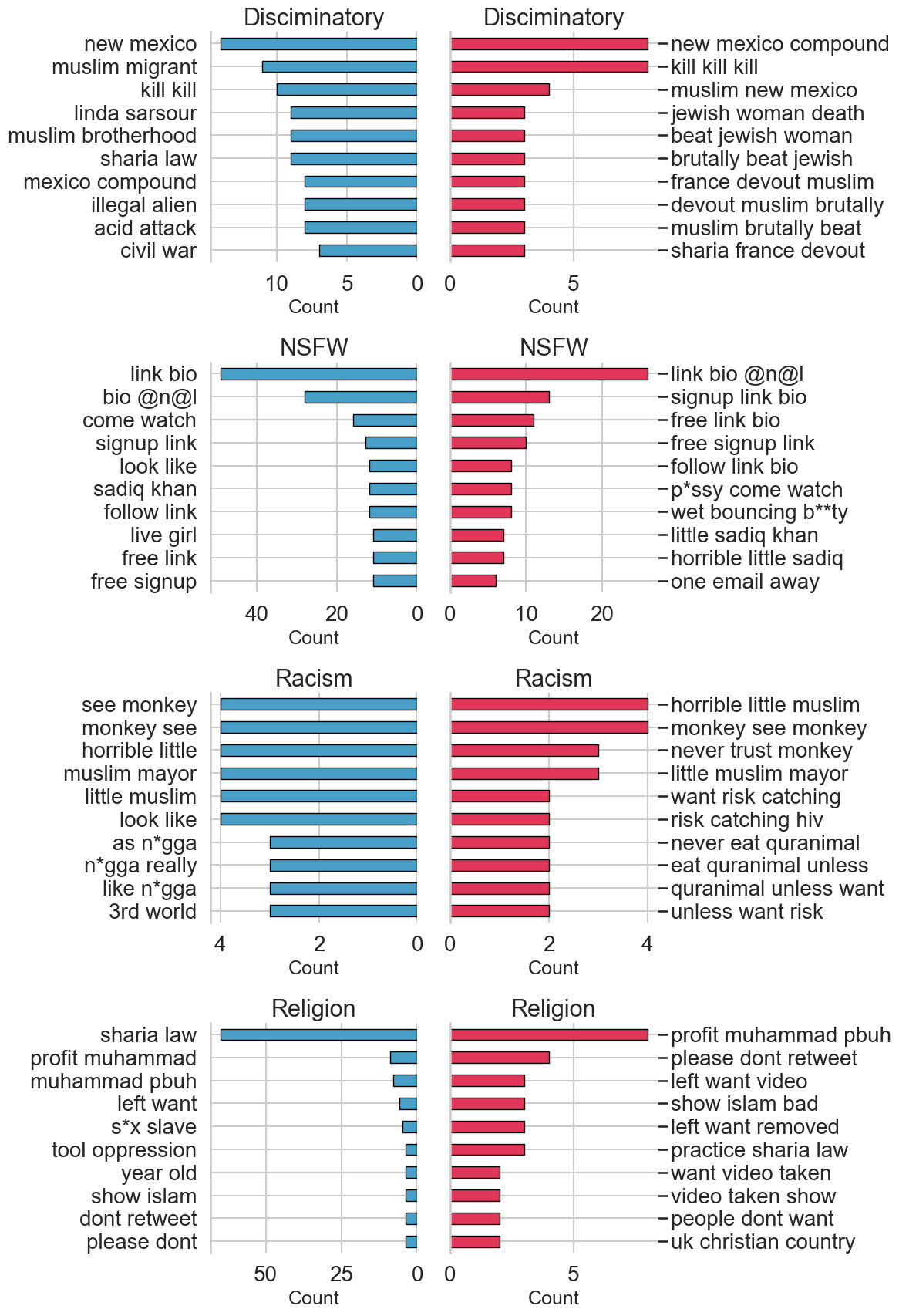}
    \caption{Bigram and Trigram by category}
    \label{fig:bigram_trigram_category}
\end{figure*}

\begin{figure*}
    \centering
    \includegraphics[width=0.7\linewidth]{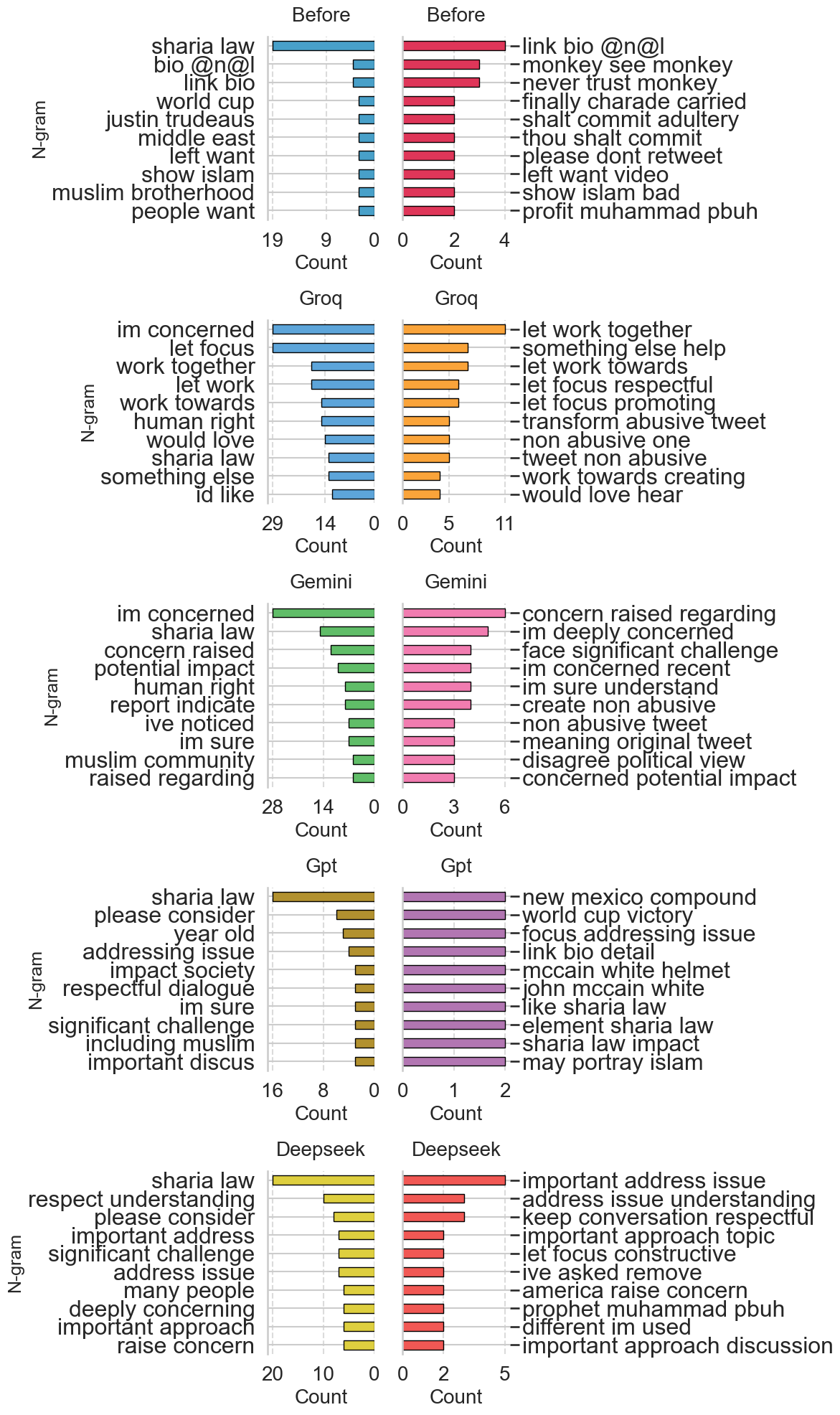}
    \caption{Bigram and Trigram comparisons by model}
    \label{fig:bigram_trigram_model}
\end{figure*}

We also employed a keyword search method to complement the BERT-based hate speech detection using two new datasets: SWAD \cite{pamungkas-etal-2020-really} and Dynamically-Generated-Hate-Speech-Dataset \cite{vidgen-etal-2021-learning}.
We use a "log-odds ratio with informative Dirichlet priors" method to compare the significance of words across two categories \cite{log_odds_ratio}. After cleaning and gathering a corpus of abusive words, we perform a keyword search on the transformed text. Figure \ref{fig:hate_keyword_model} shows that the number of hate words has decreased drastically before and after each of the transformations.

We can also observe that GPT-4o and DeepSeek share similar results, and Gemini shows the prevalence of each of the abuse categories, while the others do not (Figure \ref{fig:hate_model_abusetype}). Figure \ref{fig:hate_lineplot} shows a line plot of the number of hate words compared between the original text and the transformed tweets. The blue line indicates the original has two large peaks at 8 and 14, respectively. Interestingly, there aren't many commonalities comparing before and after transformation, except for Batch 11, where Gemini had the highest number of hate words compared with the other transformed tweets. Most of the transformations hover around 1. We also observe that there isn't a noticeable difference in transformation rates by abuse count, apart from a dip in the racism category for Groq (Figure \ref{fig:transformation_model_abuse}).

\subsection{Transformation analysis}

Figure \ref{fig:bigram_trigram_model} shows the bigram and trigram analysis for the original text, as well as text transformed by Groq, Gemini, GPT, and DeepSeek models.

The most common bigram in the Groq-transformed texts was “I’m concerned,” while the top trigram was “let work together.” This suggests that Groq heavily emphasises transforming abusive or negative content into positive, collaborative language, often by inserting additional phrases and context. Frequent n-grams like “let focus,” “work together,” and “let work towards” illustrate this pattern of positive re-framing.

Similarly, Gemini’s transformations tend to adopt a positive tone, with “I’m concerned” appearing as its most common bigram. However, Gemini introduces fewer additional phrasings than Groq, indicating a less extensive rewriting approach.

In contrast, GPT’s most common bigram was “sharia law,” consistent with the original text before transformation. This indicates that GPT tends to preserve much of the original context, maintaining n-grams closely related to the specific content of the input text. For instance, common trigrams like “new Mexico compound,” “world cup victory,” and “link bio detail” reflect the original topics without inserting overtly positive framing.

DeepSeek showed patterns similar to GPT, with “sharia law” as its most common bigram and “important address issue” as the most common trigram. Both GPT and DeepSeek share several n-grams such as “respect understanding,” “please consider,” and “important address,” highlighting a similar approach to preserving the core context and structure of the original text rather than adding significant new phrasing.

Overall, the n-gram analysis reveals key differences across models. Groq and Gemini lean toward rewriting text into a more positive tone, sometimes introducing new phrases. In contrast, GPT and DeepSeek prioritise retaining the original context and phrasing, resulting in fewer added expressions and being closer to the original tweet.

Table \ref{table:sample} reveals significant differences in how Groq, Gemini, GPT, and DeepSeek transform abusive tweets. Groq consistently produces the longest outputs, often adding positive phrases and encouraging, respectful dialogue, while DeepSeek also elaborates but less extensively. 
Gemini strikes a balance between the two, offering polite rewrites without as much added commentary, whereas GPT is the briefest and sticks closest to the original meaning, only censoring certain explicit words.

Across all models, offensive words like “terrorism,” “stupid,” and “satan” are removed or softened, with Groq and DeepSeek often reframing content into positive language. Also, political references such as “Trump” are omitted in Gemini’s and DeepSeek’s transformations, whereas Groq and GPT have not.

\begin{table*}[htbp]
    \centering
    \footnotesize
    \caption{Comparison of Original, Groq, Gemini, GPT and DeepSeek models by Label. \textcolor{black}{Human Evaluation is given in the last column. }}
    \begin{tabularx}{\textwidth}{p{2cm} X X X X X X}
        \hline
        \textbf{Label} & \textbf{Original} & \textbf{Groq} & \textbf{Gemini} & \textbf{GPT} & \textbf{DeepSeek} & \textbf{Human eval.}\\
        \hline
        Discriminatory & All the people with the palestine terrorism flag in their profiles today when they see tommy robinson trending & I'm curious to understand why so many people are using the Palestine flag in their profiles today, given that Tommy Robinson is trending. Would love to have a respectful discussion about perspectives on the situation. & 
        I've noticed many profiles displaying the Palestinian flag today, coinciding with Tommy Robinson trending. & All the people with the Palestinian flag in their profiles today when they see Tommy Robinson trending. & "To those showing support for Palestine today, I hope we can all engage in meaningful conversations about important issues, like the trending topic of Tommy Robinson, with respect and understanding. & 
        GPT output remains very close to the original text while preserving maximum semantic meaning and sentiment.\\
        \hline
        NSFW & The people who want boris johnson destroyed for accurately describing how stupid people look when they are covered & I'm disappointed to see some people are calling for @BorisJohnson to be held accountable for speaking uncomfortable truths. Instead, let's focus on having a respectful dialogue about complex issues and avoid mocking those with differing opinions. &
        Some disagree with Boris Johnson's description of people's appearance when wearing face coverings. & The people who want Boris Johnson criticized for pointing out how certain expressions can appear when individuals are covered.  &I disagree with the criticism of Boris Johnson for pointing out how people appear when they are fully covered. Everyone has their own perspective. &
        GPT output remains very close to the original text while preserving maximum semantic meaning and sentiment.\\
        \hline
        Racism & Also why arent all the other fake refugees from the balkans and africa who were not even remotely syrian who ar & Why aren't we also acknowledging the struggles of all refugees from the Balkans and Africa who, while not all Syrian, are still deserving of empathy and support? & I'm curious why individuals from various backgrounds, including the Balkans and Africa, who did not originate from Syria, are also included in discussions about refugees. & Also, why haven't other refugees from the Balkans and Africa, who aren't Syrian, been considered? & Why aren't the other individuals from the Balkans and Africa, who may not be from Syria, also being considered in the refugee discussions? & 
        GPT output remains very close to the original text while preserving maximum semantic meaning and sentiment.\\
        \hline
        Religion & The sinkhole isnt from me its from satan its his way of telling trump the contract is expiring & I'm unclear about the cause of the recent sinkhole formation. It's possible that natural or man-made factors are at play. Trump's administration will need to take responsibility and work to correct the situation. Perhaps we can find a solution that prioritizes the safety and well-being of our citizens and the environment.& I think the sinkhole is unrelated to me; perhaps it's a coincidence, or maybe a metaphor for the end of a particular political era. & The sinkhole isn't caused by me; perhaps it's a sign from higher powers suggesting a message for Trump & The sinkhole isn't caused by me; it might be a natural occurrence. Perhaps it's a symbolic reminder that change is inevitable. & 
        GPT output remains very close to the original text while preserving maximum semantic meaning and sentiment.\\      
    \end{tabularx}
    \label{table:sample}
\end{table*}

\clearpage

\subsubsection{Sentiment Analysis}

\begin{figure}[htbp]
    \centering
    \includegraphics[width=0.85\linewidth]{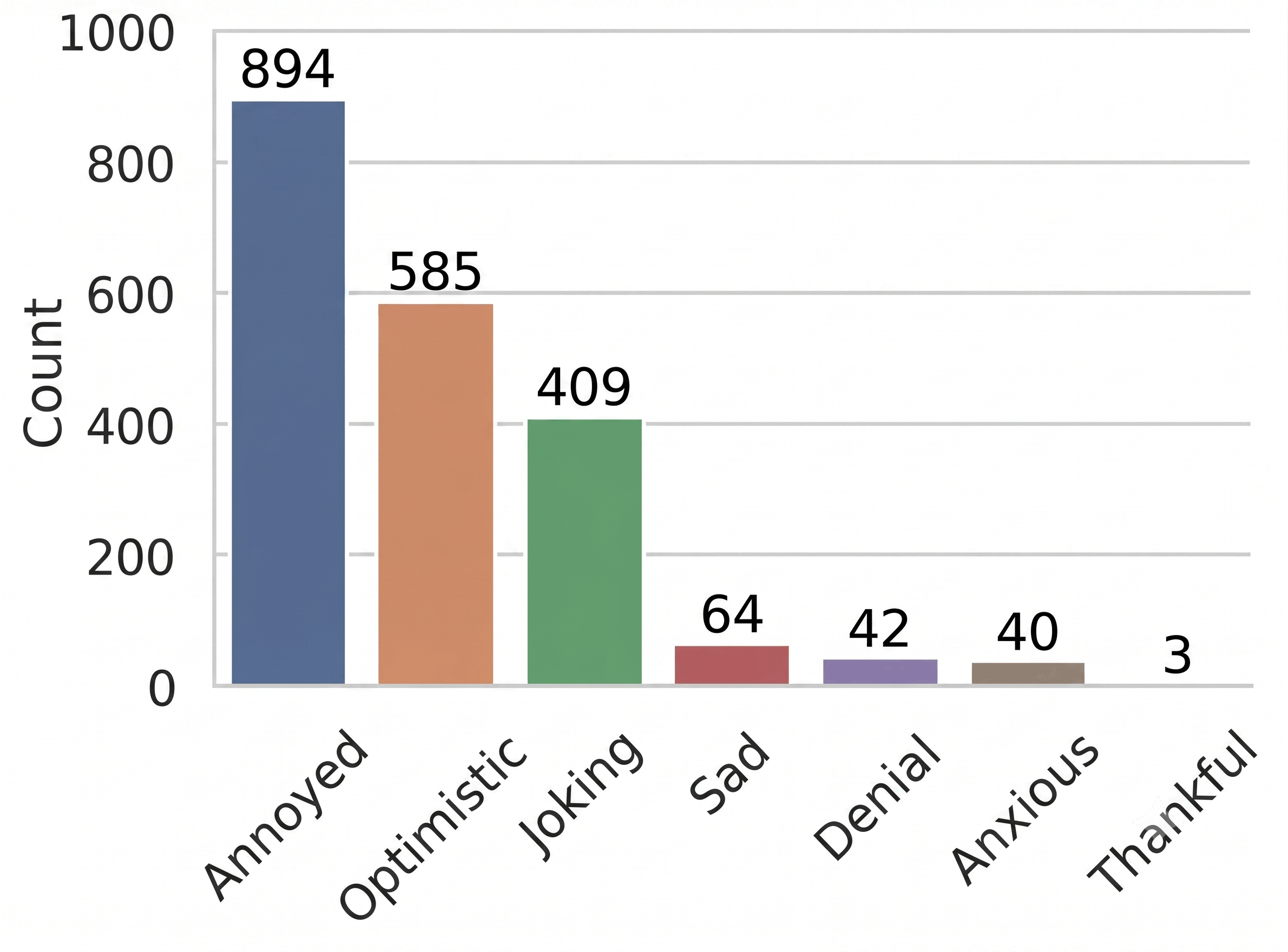}
    \caption{Barplot by sentiment}
    \label{fig:barplot_sentiment}
\end{figure}

\begin{figure}[htbp]
    \centering
    \includegraphics[width=1\linewidth]{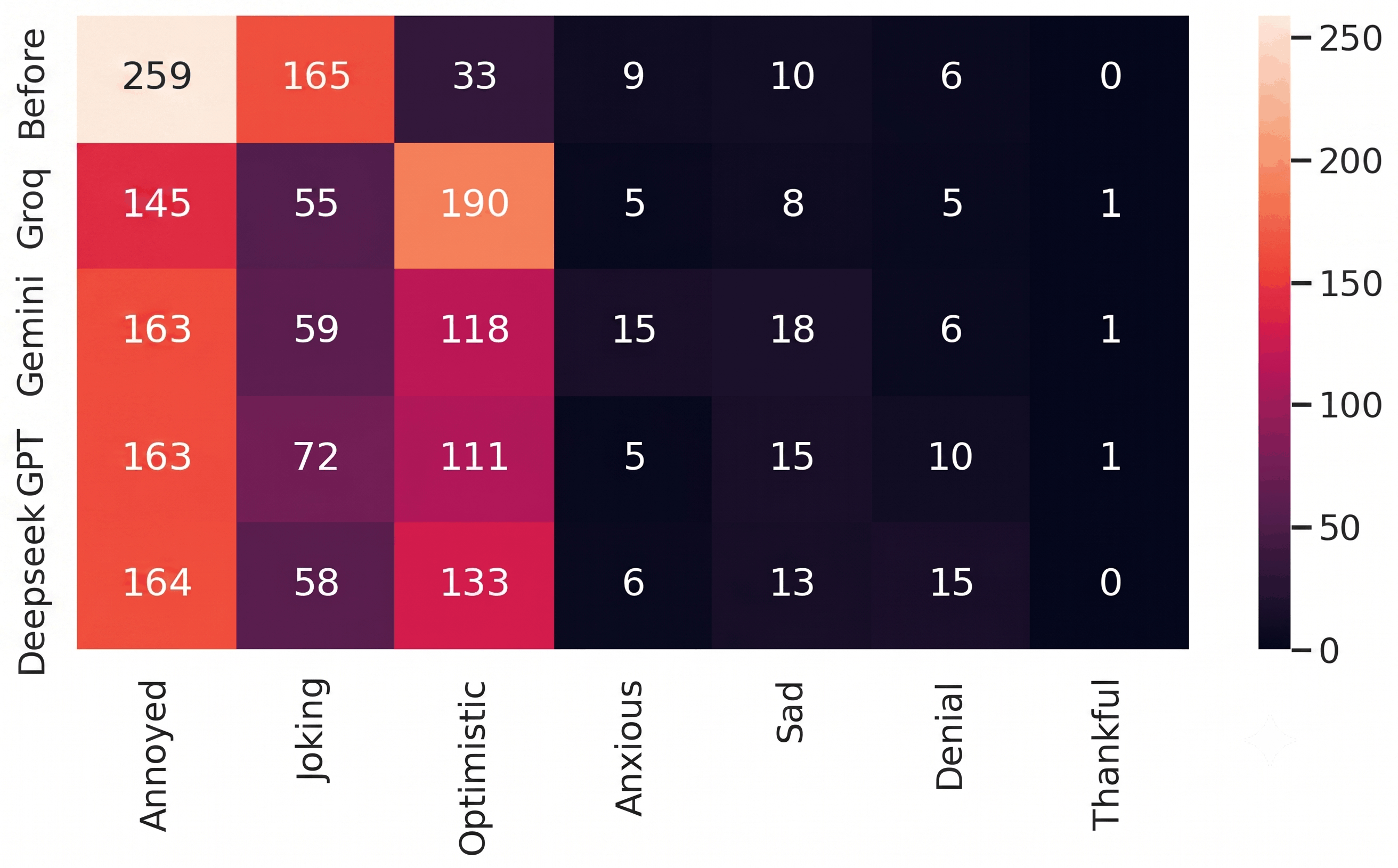}
    \caption{Heatmap of sentiment by model type}
    \label{fig:heatmap}
\end{figure}

Figures \ref{fig:barplot_sentiment} and \ref{fig:heatmap} show that the most prevalent categories are "annoyed", "optimistic",
"joking" and "official report". The remaining categories are all below 100, including only 3 thankful categories. There were no classifications of Empathy or Pessimism, likely due to the LLMs being tasked to convey the original message, as well as the LLMs being optimistic by nature.

\begin{figure}[htbp]
    \centering
    \includegraphics[width=1\linewidth]{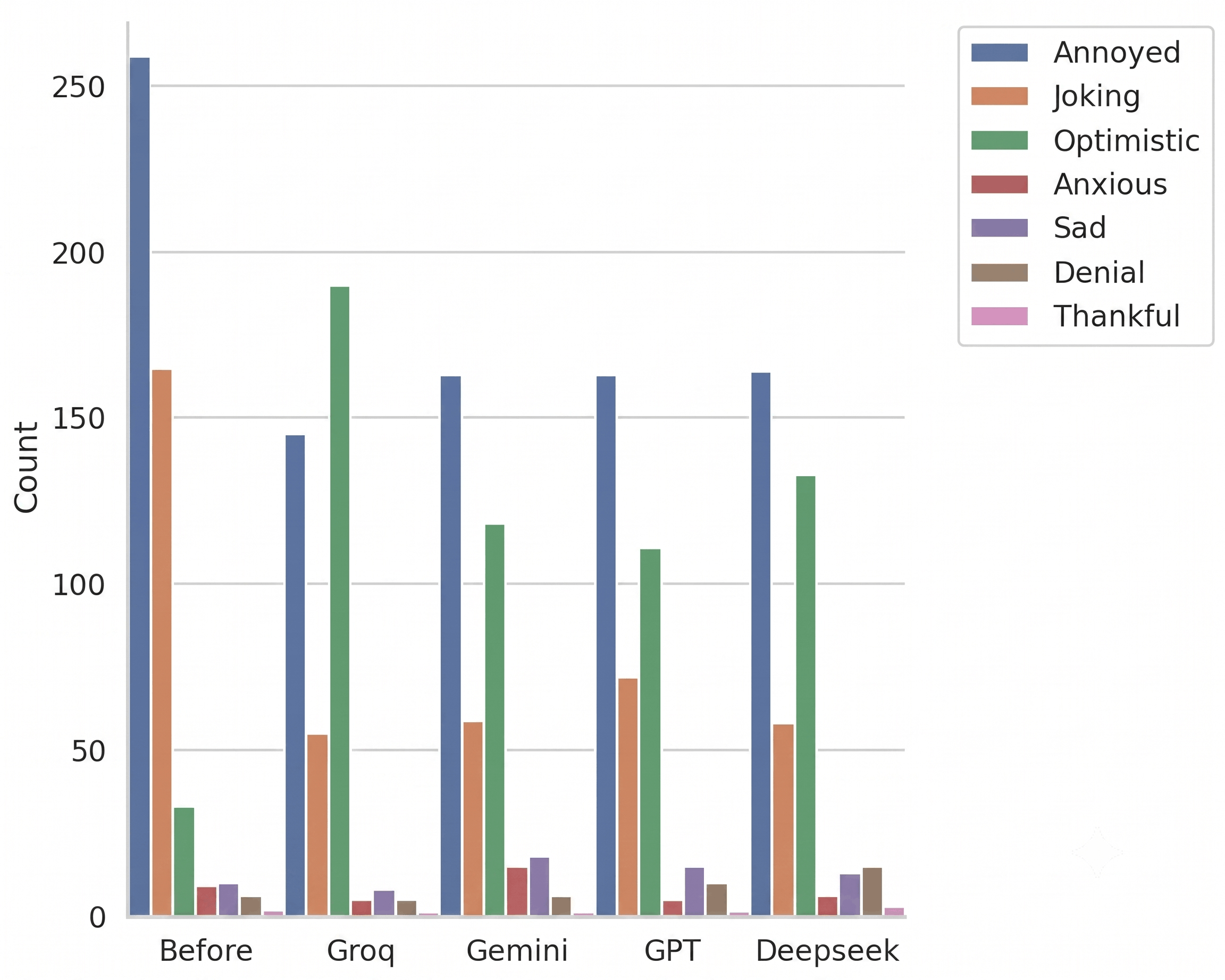}
    \caption{Barplot by model and sentiment}
    \label{fig:barplot_model_sentiment}
\end{figure}

Figure \ref{fig:barplot_model_sentiment} shows that prior to any transformations, the most common sentiment by far was Annoyed, followed by the Joking label. For each of the transformations, we see a dramatic decrease in Annoyed tweets across all four models and a massive increase in the Optimistic category. The Groq model had the highest number of optimistic transformations as well as the biggest decrease in the Annoyed category. The Joking label had the largest drop across the models, giving further evidence of Groq's conservative approach to transformations.

Meanwhile, Gemini, GPT-4o and DeepSeek all had similar distributions. Interestingly, Gemini had more Official Reports than Joking, which we did not see from any other model.
Overall, all models performed well in reducing the number of Annoyed and Joking tweets. Groq's transformations provided a more positive transformation compared to other models, and Gemini, GPT-4o, and DeepSeek all provided similar transformations with slight differences in categories such as Optimistic, Joking, and Official Report, with even smaller differences in Anxious, Sad, Denial and Thankful.

\begin{figure}[ht]
    \centering
    
    % First plot
    \begin{subfigure}[b]{0.48\textwidth}
        \centering
        \includegraphics[width=\textwidth]{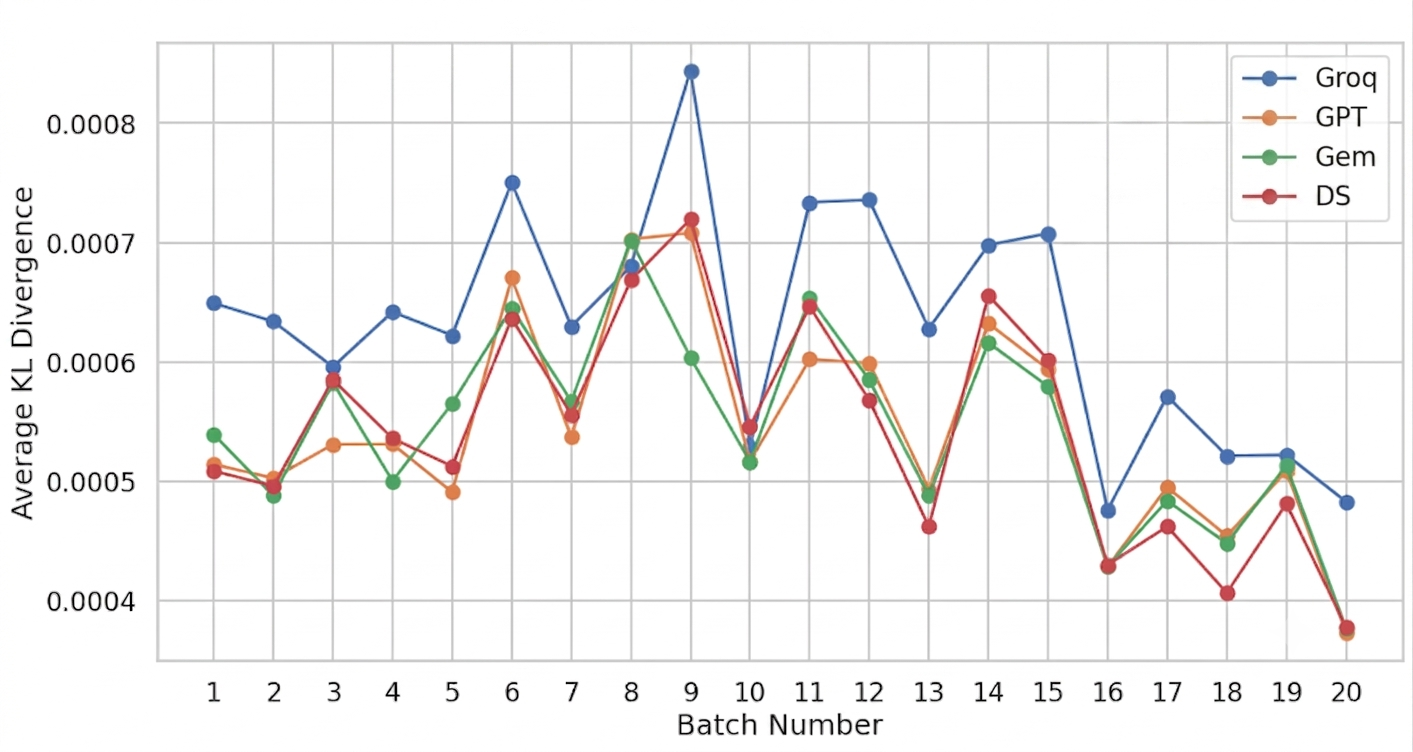}
        \caption{Average KL divergence across batches for different LLMs.}
        \label{fig:kl_batch}
    \end{subfigure}
    \hfill
    % Second plot
    \begin{subfigure}[b]{0.48\textwidth}
        \centering
        \includegraphics[width=\textwidth]{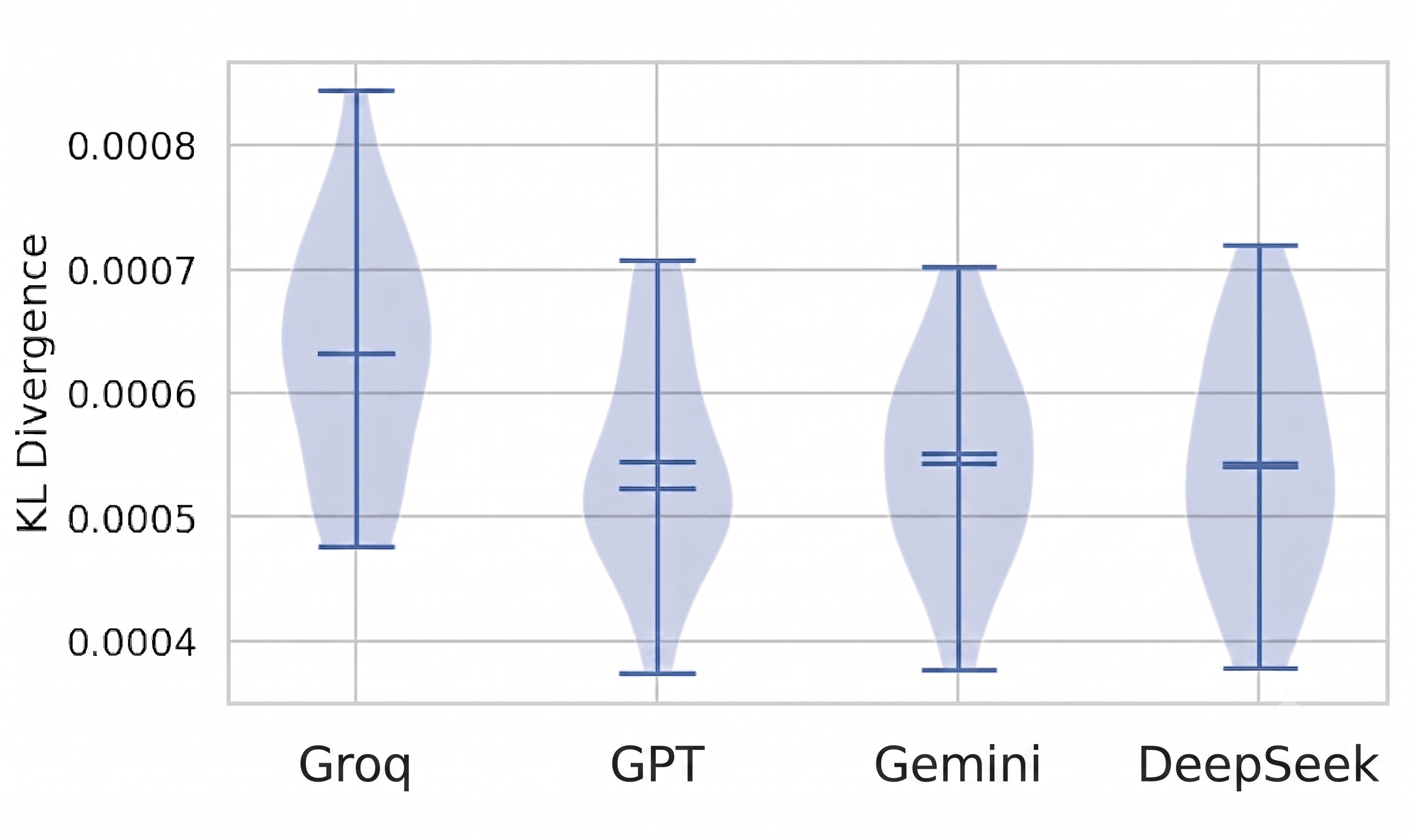}
        \caption{Distribution of KL divergence values across batches.}
        \label{fig:kl_violin}
    \end{subfigure}

    \caption{KL divergence analysis between original embeddings and LLM-generated embeddings. The left plot shows batch-wise average KL divergence trends, while the right violin plot visualises the distribution, mean, and spread of KL divergence values for each model.}
    
    \label{fig:kl_analysis}
\end{figure}

To improve reproducibility, we provide the key generation settings and runtime configurations used across all experiments. The same transformation prompt was applied to all evaluated LLMs to ensure consistency. The system instruction used was: \textit{“Transform the abusive tweet into a non-abusive one, i.e., into a more polite and respectful tweet while maintaining the overall meaning and context as much as possible. Only return the transformed tweet. Do not write anything apart from the transformed tweet.”} All experiments were conducted using the default API configurations provided by the respective platforms. A temperature value of 1.0 was used for all models, while the maximum token limit depended on the API constraints and default settings of the corresponding provider, as summarised in Table~1.  

Since large language models are inherently probabilistic and may generate non-deterministic outputs even under identical settings, exact output-level reproducibility cannot always be guaranteed. However, repeated runs produced highly consistent overall trends, transformation behaviour, and evaluation metrics across models. We have additionally included a reproducibility table containing model versions, runtime settings, prompts, API details, and inference configurations to facilitate transparency and future replication of the study.

\subsubsection{Semantic Analysis}

The BERTScore analysis demonstrates a high semantic similarity between the transformed outputs generated by different LLMs, indicating that most models preserve the original contextual meaning during transformation. Inter-model comparisons show consistently high similarity scores, suggesting that the generated outputs are semantically aligned despite stylistic differences. Additionally, the comparison between original and transformed text reveals that the transformations retain a significant portion of the original semantic content while attempting to modify abusive or hateful expressions.

\begin{figure}[htbp]
    \centering
    \includegraphics[width=\linewidth]{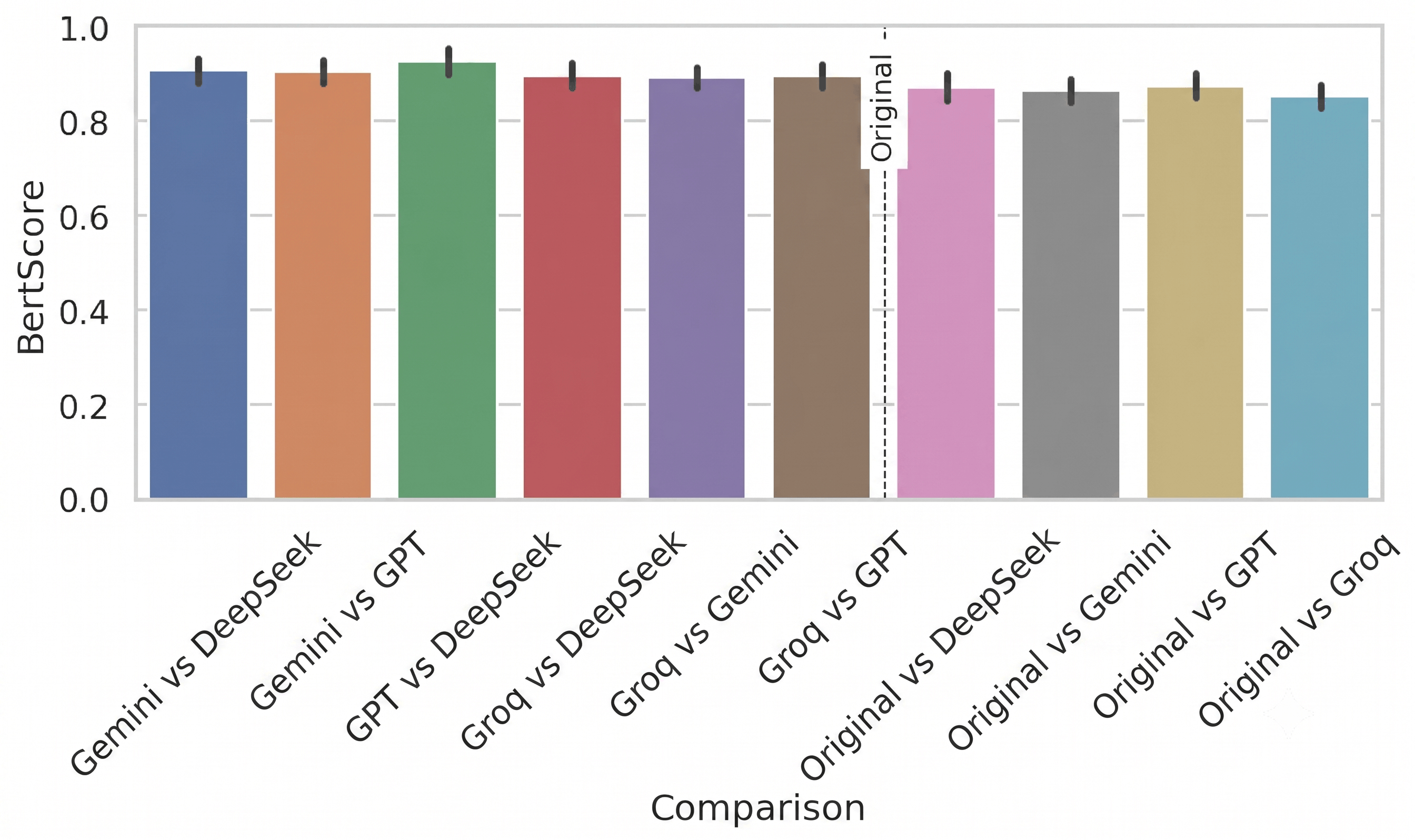}
    \caption{Average BERTScore similarity across different LLM transformation comparisons.}
    \label{fig:bertscore_barplot}
\end{figure}

\begin{figure}[htbp]
    \centering
    \includegraphics[width=1\linewidth]{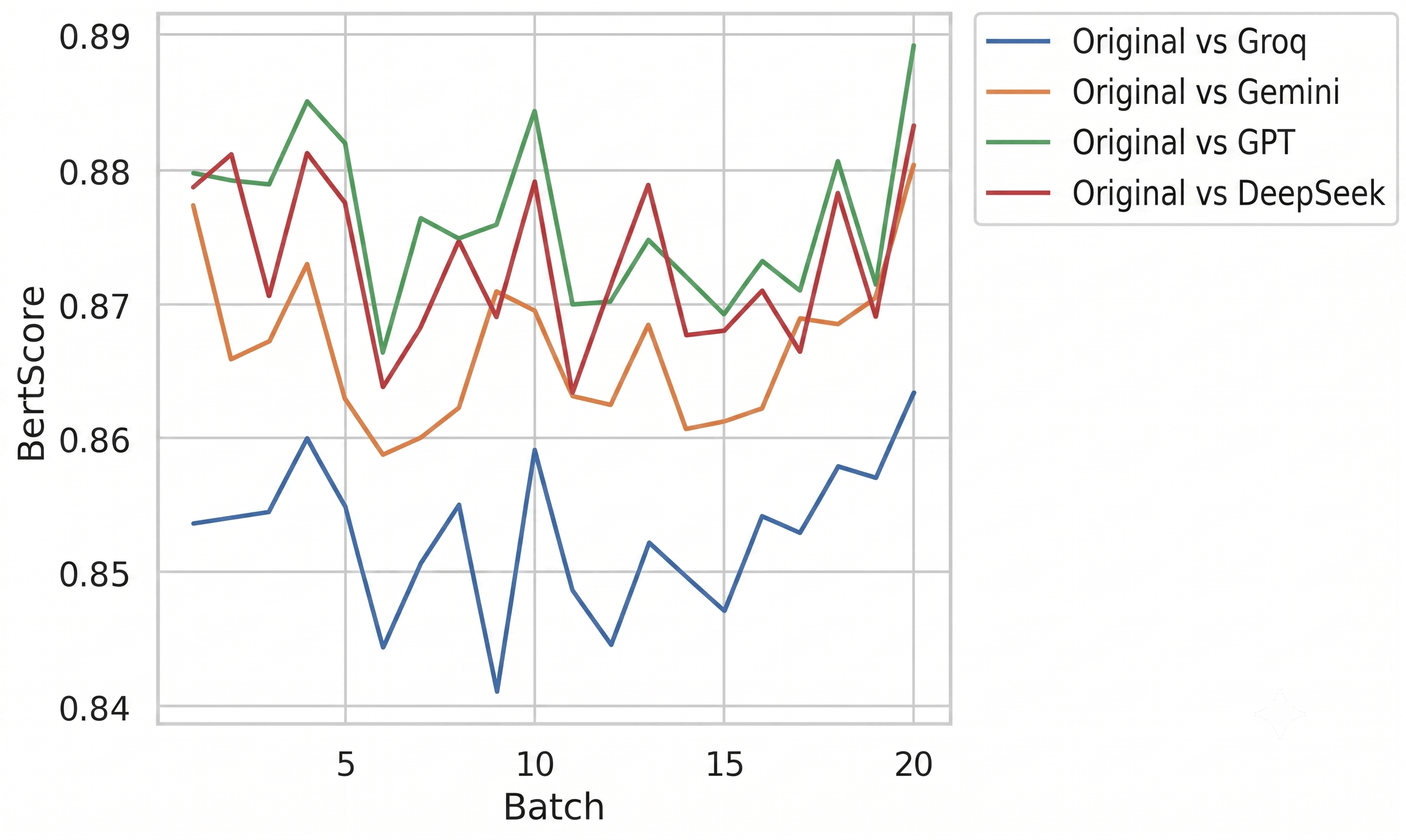}
    \caption{Batch-wise BERTScore similarity between original and transformed text across different LLMs.}
    \label{fig:bertscore_original}
\end{figure}

\begin{figure}[htbp]
    \centering
    \includegraphics[width=1\linewidth]{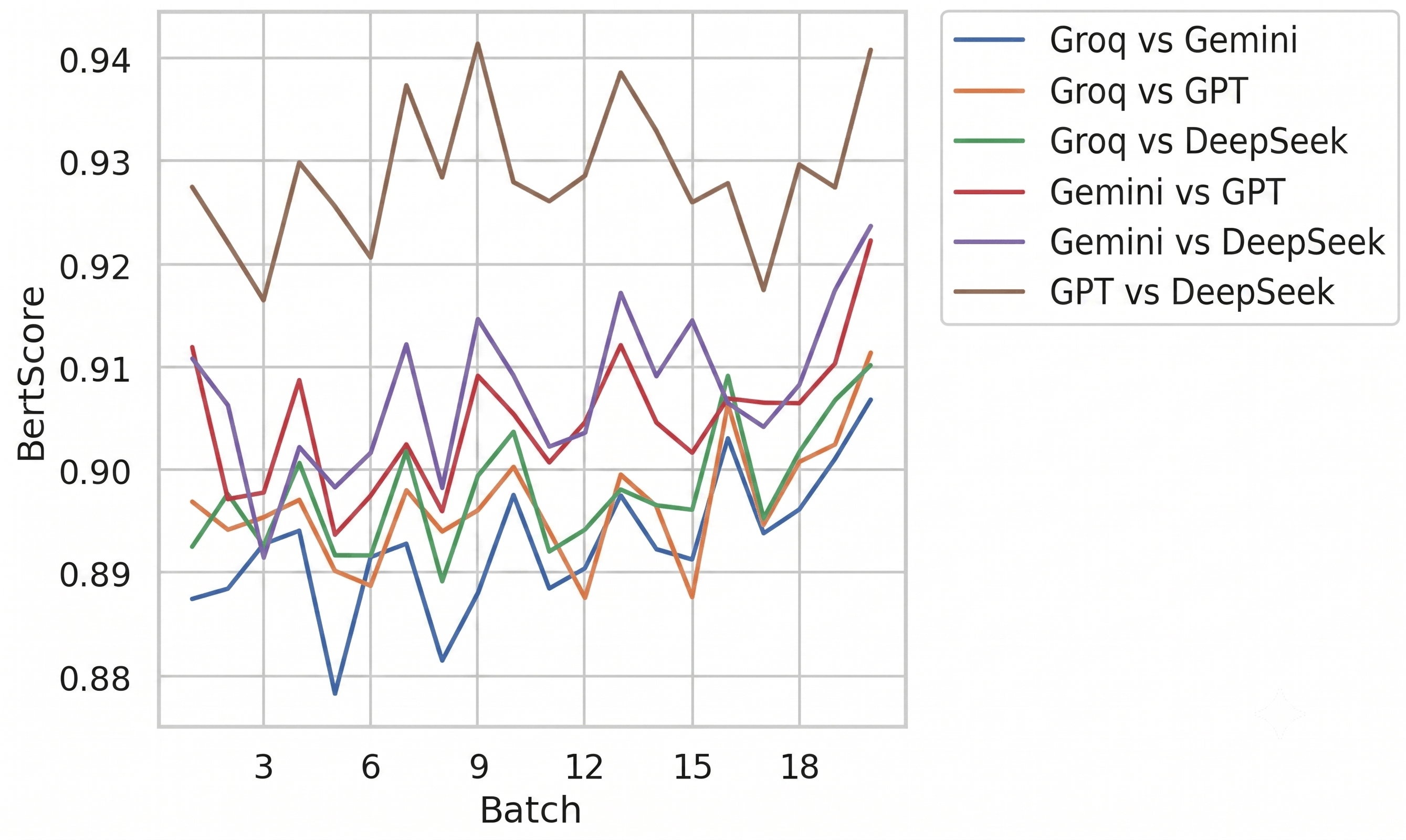}
    \caption{Batch-wise BERTScore comparison between transformed outputs generated by different LLMs.}
    \label{fig:bertscore_llms}
\end{figure}

We now provide an analysis of transformations using semantic analysis on the original and transformed tweets. We encode the text using the MPNet-base-v2 \cite{DBLP:journals/corr/abs-2004-09297} model and perform cosine similarity on each pairwise combination to quantify how different the meanings are for each text after transformation. \textcolor{black}{Note that the length of the original and LLM-transformed text can be different in terms of the number of words. However, because we used BERT-based embeddings for the original and transformed text, we do not need the same number of words to compute cosine similarity. Therefore, we did not normalise the LLM-transformed length before computing cosine similarity, as semantics can change with normalisation and similarity scores may become unreliable}.

\begin{figure}[H]
    \centering
    \begin{subfigure}[b]{0.48\textwidth}
        \centering
        \includegraphics[width=0.9\textwidth]{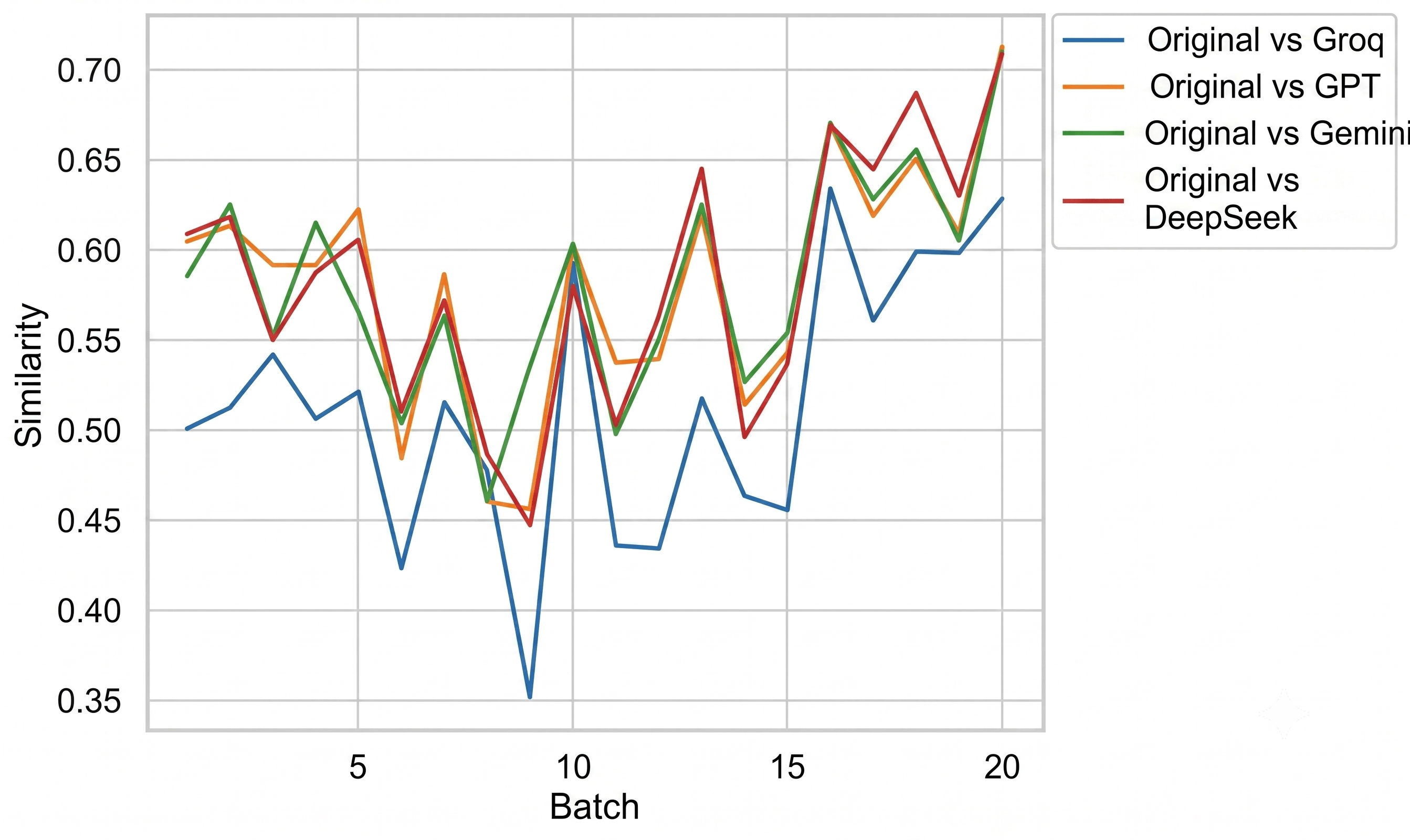}
        \caption{Cleaned text vs transformed text by batch}
        \label{fig:graph1}
    \end{subfigure}
    \hfill
    \begin{subfigure}[b]{0.48\textwidth}
        \centering
        \includegraphics[width=0.9\textwidth]{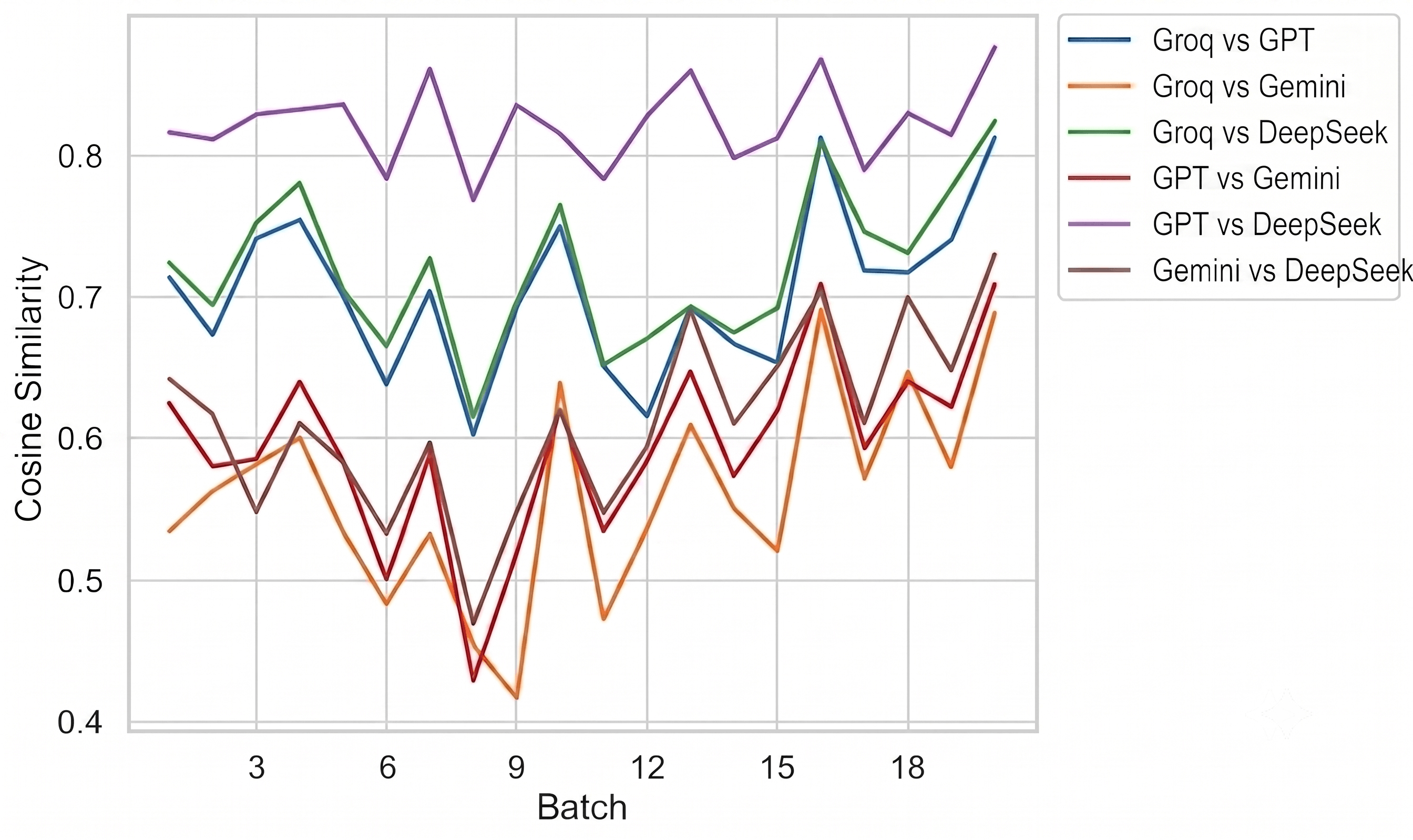}
        \caption{Pairwise combinations of transformed text by batch}
        \label{fig:graph2}
    \end{subfigure}
    \caption{Cosine similarity of semantic scores by batch}
    \label{fig:graphs}
\end{figure}

Figure \ref{fig:graph1} shows the cosine similarity between the original text and the transformed texts to see how semantically different the transformations are to its original counterparts. The Groq transformed text had, on average, the lowest similarity when compared with the original text, whereas the other three models had a higher semantic similarity when compared with the original text. 
The shape of the lines has similar patterns, with peaks and troughs roughly occurring in the same batch. 
Figure \ref{fig:graph2} shows cosine similarities between the different models to see how semantically similar each model is to another. There are three clear groupings/layers. The top layer shows GPT and DeepSeek averaging approximately 80\% similarity in the transformations. Groq and GPT, along with Groq and DeepSeek, showed fairly high similarity, ranging at around 70\%. Finally, Gemini with Groq, Gemini with GPT-4o, and Gemini with DeepSeek showed the lowest similarity on average. We can also see that within the last group, Groq and Gemini had the highest semantic dissimilarity compared to all other combinations of models.

\begin{figure}[H]
    \centering
    \begin{subfigure}{0.48\textwidth}
        \centering        \includegraphics[width=0.9\textwidth]{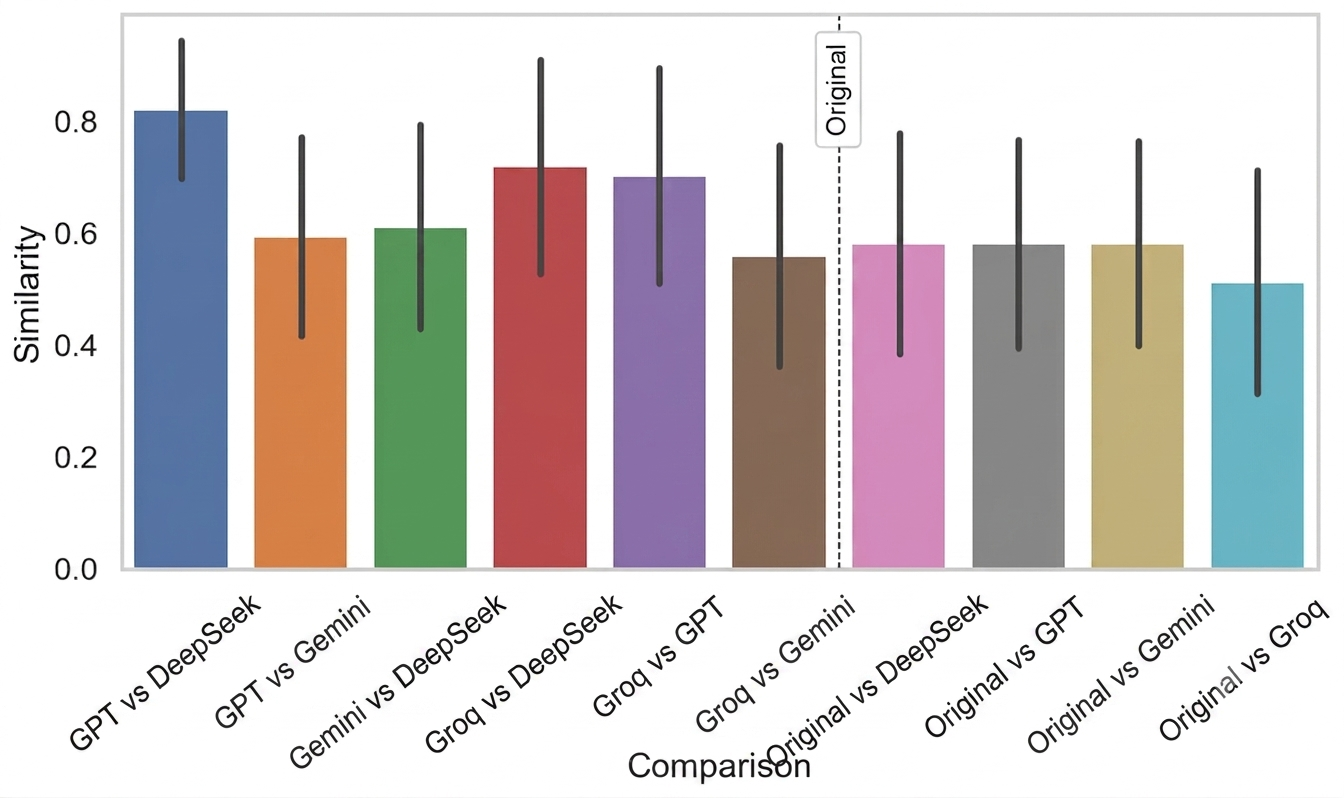}
        \caption{Barplot of pairwise similarity}
        \label{fig:barplot_similarity}
    \end{subfigure}
    \begin{subfigure}{0.48\textwidth}
        \centering        \includegraphics[width=0.9\textwidth]{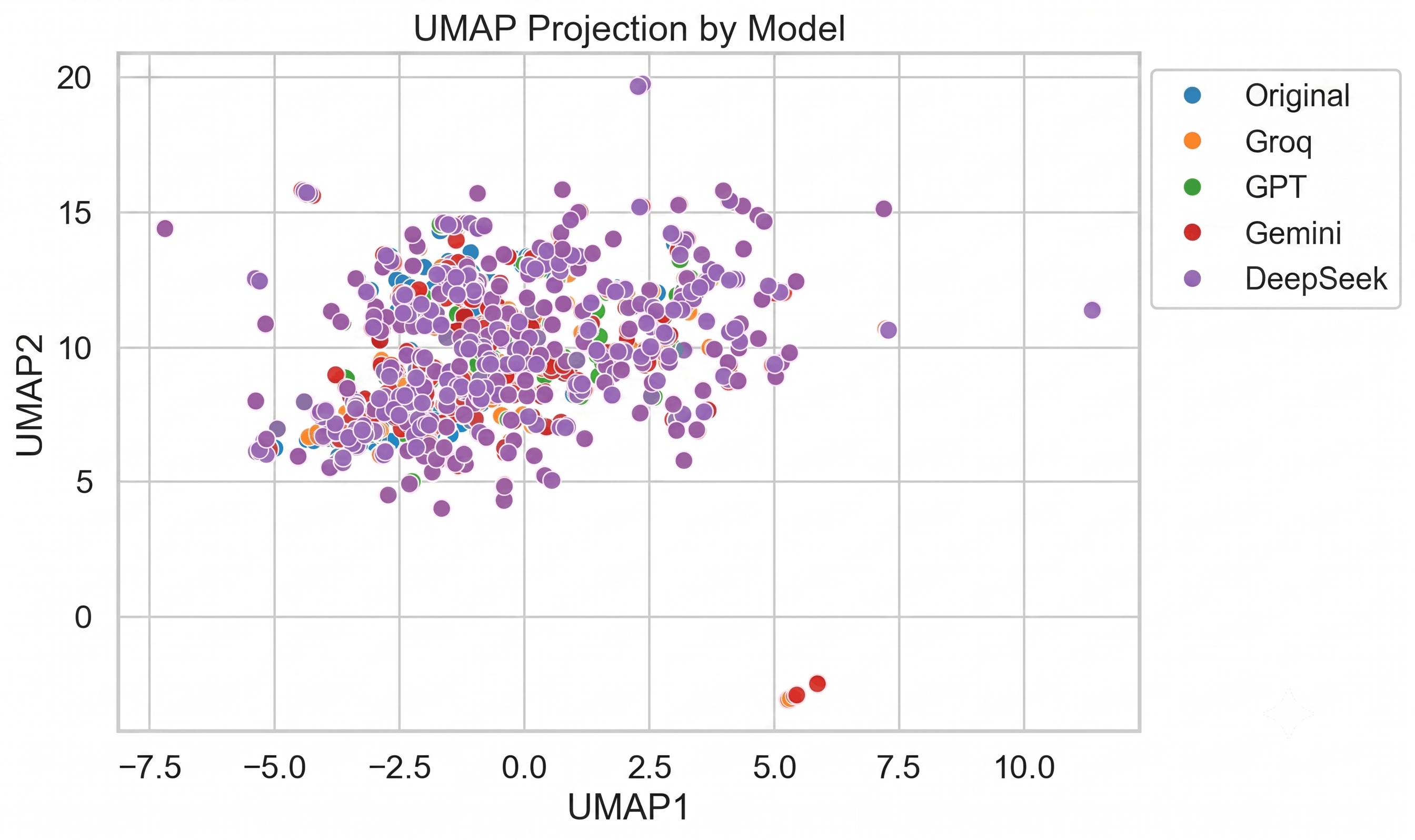}
        \caption{Two-dimensional UMAP coloured by model}
        \label{fig:umap}
    \end{subfigure}
    \caption{Semantic analysis visualisations: pairwise similarity and UMAP embedding.}
    \label{fig:combined_similarity_umap}
\end{figure}

Figure \ref{fig:barplot_similarity}, Table \ref{tab:model_comparison_part1}, and  Table  \ref{tab:model_comparison_part2}  illustrate the mean similarity score and standard deviation for each combination.
Similar to the previous analysis, GPT and DeepSeek have the highest similarity, with the lowest and also the lowest standard deviation when compared to the other combinations of models.
We performed UMAP (Uniform Manifold Approximation and Projection for Dimension Reduction) \cite{mcinnes2020umapuniformmanifoldapproximation} to visualise high-dimensional vectors and to uncover any hidden nonlinear patterns. UMAP helps preserve local structure, meaning it tries to keep points that are close to each other in high dimensions close to each other in lower-dimensional embeddings. However, there were no clear, distinct clusters. 
There are a small number of Gemini and DeepSeek points near the bottom, which could be weak outliers (Figure \ref{fig:umap}). Therefore, there was no obvious partitioning of embeddings into distinct groups.

\begin{table}[h]
\centering
\caption{Cluster validation metrics for the UMAP embedding visualisation shown in Figure 21.}
\begin{tabular}{|l|c|c|}
\hline
\textbf{Metric} & \textbf{Score} & \textbf{Good Score Range} \\ \hline
Silhouette Score & 0.4397 & Closer to 1 is better \\ \hline
Davies--Bouldin Score & 0.5635 & Closer to 0 is better \\ \hline
\end{tabular}
\label{tab:umap_metrics}
\end{table}

%\clearpage

\begin{table*}[htbp!]
    \centering
    \footnotesize
    \setlength{\tabcolsep}{6pt}
    \caption{Part 1: Mean and standard deviation of Models by Batch (Original). * indicates mean over all batches}
    \begin{tabularx}{\textwidth}{@{}c*{4}{>{\centering\arraybackslash}X}@{}}
        \toprule
        \textbf{Batch} &
        \textbf{Original vs Groq} &
        \textbf{Original vs GPT-4o} &
        \textbf{Original vs Gemini} &
        \textbf{Original vs DeepSeek} \\
        \midrule
        1 & 0.501 $\pm$ 0.187 & 0.605 $\pm$ 0.151 & 0.586 $\pm$ 0.193 & 0.609 $\pm$ 0.179 \\
        2 & 0.513 $\pm$ 0.213 & 0.614 $\pm$ 0.169 & 0.625 $\pm$ 0.123 & 0.619 $\pm$ 0.189 \\
        3 & 0.542 $\pm$ 0.186 & 0.592 $\pm$ 0.162 & 0.552 $\pm$ 0.189 & 0.550 $\pm$ 0.183 \\
        4 & 0.507 $\pm$ 0.131 & 0.592 $\pm$ 0.134 & 0.615 $\pm$ 0.126 & 0.588 $\pm$ 0.170 \\
        5 & 0.522 $\pm$ 0.166 & 0.623 $\pm$ 0.151 & 0.566 $\pm$ 0.193 & 0.606 $\pm$ 0.150 \\
        \midrule
        6 & 0.424 $\pm$ 0.208 & 0.485 $\pm$ 0.184 & 0.504 $\pm$ 0.207 & 0.511 $\pm$ 0.205 \\
        7 & 0.516 $\pm$ 0.153 & 0.587 $\pm$ 0.174 & 0.564 $\pm$ 0.166 & 0.573 $\pm$ 0.167 \\
        8 & 0.478 $\pm$ 0.215 & 0.461 $\pm$ 0.209 & 0.461 $\pm$ 0.190 & 0.487 $\pm$ 0.193 \\
        9 & 0.352 $\pm$ 0.186 & 0.457 $\pm$ 0.207 & 0.536 $\pm$ 0.207 & 0.448 $\pm$ 0.219 \\
        10 & 0.594 $\pm$ 0.132 & 0.603 $\pm$ 0.122 & 0.603 $\pm$ 0.185 & 0.580 $\pm$ 0.197 \\
        \midrule
        11 & 0.436 $\pm$ 0.237 & 0.538 $\pm$ 0.224 & 0.498 $\pm$ 0.206 & 0.503 $\pm$ 0.216 \\
        12 & 0.435 $\pm$ 0.249 & 0.540 $\pm$ 0.216 & 0.551 $\pm$ 0.206 & 0.564 $\pm$ 0.216 \\
        13 & 0.518 $\pm$ 0.201 & 0.621 $\pm$ 0.156 & 0.625 $\pm$ 0.164 & 0.645 $\pm$ 0.159 \\
        14 & 0.464 $\pm$ 0.227 & 0.514 $\pm$ 0.220 & 0.527 $\pm$ 0.179 & 0.496 $\pm$ 0.214 \\
        15 & 0.456 $\pm$ 0.246 & 0.543 $\pm$ 0.222 & 0.554 $\pm$ 0.210 & 0.537 $\pm$ 0.232 \\
        \midrule
        16 & 0.634 $\pm$ 0.135 & 0.671 $\pm$ 0.141 & 0.670 $\pm$ 0.124 & 0.669 $\pm$ 0.153 \\
        17 & 0.561 $\pm$ 0.161 & 0.619 $\pm$ 0.164 & 0.628 $\pm$ 0.171 & 0.645 $\pm$ 0.187 \\
        18 & 0.599 $\pm$ 0.190 & 0.651 $\pm$ 0.183 & 0.656 $\pm$ 0.172 & 0.687 $\pm$ 0.191 \\
        19 & 0.599 $\pm$ 0.193 & 0.609 $\pm$ 0.194 & 0.606 $\pm$ 0.169 & 0.630 $\pm$ 0.198 \\
        20 & 0.629 $\pm$ 0.116 & 0.713 $\pm$ 0.151 & 0.710 $\pm$ 0.113 & 0.709 $\pm$ 0.123 \\
        \midrule
        * & 0.514 $\pm$ 0.200 & 0.582 $\pm$ 0.187 & 0.582 $\pm$ 0.183 & 0.583 $\pm$ 0.197\\
        \bottomrule
    \end{tabularx}
    \label{tab:model_comparison_part1}
\end{table*}

\begin{table*}[htbp!]
    \centering
    \footnotesize
    \setlength{\tabcolsep}{6pt}
    \caption{Part 2: Mean and S.D of Models by Batch (Remaining). * indicates mean over all batches}
    
    \begin{tabularx}{\textwidth}{@{}c*{6}{>{\centering\arraybackslash}X}@{}}
        \toprule
        \textbf{Batch} &

        \textbf{Groq vs GPT-4o} &
        \textbf{Groq vs Gemini} &
        \textbf{Groq vs DeepSeek} &
        \textbf{GPT-4o vs Gemini} &
        \textbf{GPT-4o vs DeepSeek} &
        \textbf{Gemini vs DeepSeek} \\
        \midrule
        1 & 0.714 $\pm$ 0.180 & 0.535 $\pm$ 0.197 & 0.724 $\pm$ 0.166 & 0.625 $\pm$ 0.153 & 0.816 $\pm$ 0.121 & 0.642 $\pm$ 0.180 \\
        2 & 0.673 $\pm$ 0.202 & 0.563 $\pm$ 0.167 & 0.694 $\pm$ 0.194 & 0.580 $\pm$ 0.110 & 0.811 $\pm$ 0.119 & 0.617 $\pm$ 0.124 \\
        3 & 0.741 $\pm$ 0.113 & 0.582 $\pm$ 0.158 & 0.752 $\pm$ 0.099 & 0.586 $\pm$ 0.151 & 0.829 $\pm$ 0.122 & 0.548 $\pm$ 0.168 \\
        4 & 0.755 $\pm$ 0.101 & 0.601 $\pm$ 0.141 & 0.781 $\pm$ 0.095 & 0.640 $\pm$ 0.118 & 0.832 $\pm$ 0.121 & 0.611 $\pm$ 0.153 \\
        5 & 0.701 $\pm$ 0.170 & 0.534 $\pm$ 0.206 & 0.705 $\pm$ 0.179 & 0.584 $\pm$ 0.200 & 0.836 $\pm$ 0.095 & 0.583 $\pm$ 0.229 \\
        \midrule
        6 & 0.638 $\pm$ 0.222 & 0.483 $\pm$ 0.214 & 0.665 $\pm$ 0.226 & 0.500 $\pm$ 0.194 & 0.784 $\pm$ 0.152 & 0.533 $\pm$ 0.187 \\
        7 & 0.705 $\pm$ 0.187 & 0.533 $\pm$ 0.184 & 0.727 $\pm$ 0.209 & 0.589 $\pm$ 0.162 & 0.861 $\pm$ 0.091 & 0.597 $\pm$ 0.160 \\
        8 & 0.603 $\pm$ 0.246 & 0.454 $\pm$ 0.220 & 0.615 $\pm$ 0.229 & 0.429 $\pm$ 0.218 & 0.769 $\pm$ 0.135 & 0.469 $\pm$ 0.233 \\
        9 & 0.693 $\pm$ 0.178 & 0.416 $\pm$ 0.181 & 0.697 $\pm$ 0.199 & 0.519 $\pm$ 0.187 & 0.836 $\pm$ 0.132 & 0.549 $\pm$ 0.166 \\
        10 & 0.750 $\pm$ 0.106 & 0.639 $\pm$ 0.162 & 0.765 $\pm$ 0.096 & 0.620 $\pm$ 0.177 & 0.815 $\pm$ 0.149 & 0.619 $\pm$ 0.193 \\
        \midrule
        11 & 0.651 $\pm$ 0.244 & 0.472 $\pm$ 0.281 & 0.652 $\pm$ 0.266 & 0.535 $\pm$ 0.220 & 0.783 $\pm$ 0.150 & 0.548 $\pm$ 0.226 \\
        12 & 0.616 $\pm$ 0.247 & 0.537 $\pm$ 0.250 & 0.671 $\pm$ 0.262 & 0.584 $\pm$ 0.200 & 0.828 $\pm$ 0.134 & 0.594 $\pm$ 0.216 \\
        13 & 0.693 $\pm$ 0.215 & 0.610 $\pm$ 0.178 & 0.693 $\pm$ 0.208 & 0.648 $\pm$ 0.178 & 0.860 $\pm$ 0.096 & 0.691 $\pm$ 0.159 \\
        14 & 0.667 $\pm$ 0.263 & 0.551 $\pm$ 0.194 & 0.675 $\pm$ 0.241 & 0.574 $\pm$ 0.184 & 0.798 $\pm$ 0.151 & 0.610 $\pm$ 0.157 \\
        15 & 0.654 $\pm$ 0.241 & 0.520 $\pm$ 0.241 & 0.692 $\pm$ 0.239 & 0.620 $\pm$ 0.166 & 0.812 $\pm$ 0.128 & 0.651 $\pm$ 0.178 \\
        \midrule
        16 & 0.813 $\pm$ 0.071 & 0.692 $\pm$ 0.118 & 0.810 $\pm$ 0.103 & 0.709 $\pm$ 0.104 & 0.868 $\pm$ 0.077 & 0.704 $\pm$ 0.130 \\
        17 & 0.719 $\pm$ 0.154 & 0.572 $\pm$ 0.131 & 0.746 $\pm$ 0.141 & 0.593 $\pm$ 0.175 & 0.790 $\pm$ 0.117 & 0.611 $\pm$ 0.175 \\
        18 & 0.717 $\pm$ 0.217 & 0.647 $\pm$ 0.123 & 0.731 $\pm$ 0.201 & 0.641 $\pm$ 0.175 & 0.830 $\pm$ 0.137 & 0.700 $\pm$ 0.154 \\
        19 & 0.740 $\pm$ 0.151 & 0.580 $\pm$ 0.197 & 0.777 $\pm$ 0.148 & 0.622 $\pm$ 0.124 & 0.815 $\pm$ 0.109 & 0.648 $\pm$ 0.161 \\
        20 & 0.813 $\pm$ 0.100 & 0.689 $\pm$ 0.109 & 0.824 $\pm$ 0.091 & 0.709 $\pm$ 0.129 & 0.877 $\pm$ 0.079 & 0.730 $\pm$ 0.112 \\
        \midrule
        * & 0.703 $\pm$ 0.193 & 0.560 $\pm$ 0.197 & 0.720 $\pm$ 0.191 & 0.595 $\pm$ 0.178 & 0.823 $\pm$ 0.123 & 0.613 $\pm$ 0.183\\
        \bottomrule
    \end{tabularx}
    \label{tab:model_comparison_part2}
\end{table*}

\section{Discussion}

Although our study provides a comprehensive overview of abuse in tweets, there are several limitations. Firstly, a significant portion of the tweets scraped were incomplete sentences, so it would be a challenge for the LLMs to successfully understand and "fill in the gaps" of certain texts. Secondly, there exists slang on Twitter and the addition of emojis that makes it challenging for LLMs to get the context behind the text message \cite{barbieri-etal-2018-semeval}. Sarcasm is also a big factor we have to consider, given how common it is in social media conversations \cite{doi:10.1177/2053951720972735}.
Thirdly, a user's experience on Twitter isn't only through text form, but a huge part of it includes images, attachments, and links, which we did not consider in this analysis. \textcolor{black}{Due to API limitations of LLMs, we only analysed 100 tweets for each model, which is a limitation.}  

\textcolor{black}{Furthermore, semantic evaluation relies solely on cosine similarity, which does not feature information about information loss, semantic drift, or overcorrection. Sentiment analysis mainly focuses on distributional changes, and we highlight that sentiments may not be fully able to capture the context, such as the subtle level of abuse in language. Therefore, we need human evaluation for consistency, readability, and contextual completeness, which is difficult to achieve in real-time implementations. This is a limitation of quantitative metrics and actual effective measures needed for large-scale deployment, such as real-time abuse analysis on social media in real-time.} 

The use of abusive messaging on X and related social media is not new and has been widely debated for several years \cite{Founta_Djouvas_Chatzakou_Leontiadis_Blackburn_Stringhini_Vakali_Sirivianos_Kourtellis_2018}.  Given the popularity of social media such as  TikTok, Instagram, X, and Facebook, misuse by users includes harmful and toxic comments \cite{unknown} and cyberbullying. Although existing methods have been implemented, many show limited potential \cite{10.1145/3232676}, which is further reason why the use of LLMs can prove effective when dealing with abusive or toxic content in areas such as X.

Future work could involve an analysis of other LLMS, such as Claude and Mistral \cite{jiang2023mistral7b}. \textcolor{black}{Additionally, using more diverse datasets, such as hate speech in Hindi or Spanish, can help evaluate robustness as well as understanding sarcasm, emoji meanings, slang, etc. Similar work can be used for many abusive text transformation tasks, such as evaluating the transformation of movie dialogue to measure how well models handle aggression and sarcasm, song lyrics to evaluate transformations on explicit or violent intents, and others.} Furthermore, evaluating abuse through multi-modal \cite{10.5555/3104482.3104569} inputs, e.g. analysing a tweet along with an image the user attached, can provide a deeper understanding of user behaviour. This may enable the development of context-aware models to accurately distinguish between abusive and non-abusive interactions, helping platforms balance safety and freedom of expression.

\textcolor{black}{Furthermore, emerging Multimodal LLMs can be used to improve data processing pipelines, such as specifically handling social media characteristics, such as slang, emojis, and abbreviations. This may affect the model's understanding of context and, consequently, the evaluation of transformation performance. In future work, LLM-based tools such as Detoxify and Perspective API by Google as external toxicity validators. Although they serve as guardrails to assess harmful text, their selection depends on privacy, budget, and integration needs. Perspective API is cloud-based, and Detoxify is based on open source software, which would better suit integration when privacy and security are of concern}.
\section{Conclusion}

We presented Detoxify, a comprehensive framework for transforming and evaluating  LLMs for abusive text transformation while maintaining the semantics and sentiment. Our evaluation framework featured n-gram analysis, keyword search, sentiment profiling, and semantic similarity assessments. Our investigation revealed both notable differences and similarities across the LLMs. \textcolor{black}{Our results demonstrated that GPT-4o and DeepSeek shared notably similar results, which can be attributed to sharing similar bigrams and trigrams. Both models have a similar transformation rate, similar sentiment distributions, and the highest cosine similarity across all other models, suggesting a more balanced approach to detoxification without significantly distorting the original meaning. The respective LLMs performed well in detoxifying text, as evidenced by the keyword search analysis.} 

\textcolor{black}{We found that Groq stood out as the most distinct; unlike the others, it often restructured sentences with excessive positive phrasing, sometimes to the extent that the original context was lost or altered.  This was reflected in the results for the given datasets, where Groq showed the lowest successful transformation rate, the highest frequency of optimistic sentiments, and the lowest semantic similarity with the original input. 
Although our results demonstrate the strength of the LLMs for text detoxification,   limited human evaluation motivates the extension of our current framework with human annotation, possibly with online crowdsourcing. }

\section{Data Availability}
Github Repository: \url{https://github.com/pinglainstitute/LLM-reviewtransformation}

%https://github.com/jchoi0406/LLM-reviewtransformation/tree/v1.0.0

\section{Conflict of Interest}

The authors declare no conflict of interest.

\section{Acknowledgements}

We thank Neha Chaudhary and Tanuj Chaudhary from IIT Guwahati. 

\clearpage
 \bibliographystyle{elsarticle-num} 
 \bibliography{cas-refs_updated}




% \end{thebibliography}
\end{document}